\documentclass{article}

\usepackage[
  style=numeric,
  backend=biber]{biblatex}

\usepackage{siunitx}
\usepackage{amsthm}
\usepackage{amsmath}
\usepackage{amssymb}
\usepackage{algorithm}
\usepackage{algpseudocode}
\usepackage[beginLComment=/*~,endLComment=~*/]{algpseudocodex}
\usepackage{mathtools}
\usepackage{float}
\usepackage{tabularx}
\usepackage{subfigure}
\usepackage{graphics}
\usepackage{graphicx}
\usepackage{pgfplots}
\usepackage{listings}
\usepackage{hyperref}
\usepackage{caption}
\usepackage{subcaption} 
\pgfplotsset{compat=1.18}

\DeclareMathOperator*{\argmin}{arg\,min}

\usepackage[most]{tcolorbox}
\usepackage{xcolor}

\definecolor{codegreen}{rgb}{0,0.6,0}
\definecolor{codegray}{rgb}{0.5,0.5,0.5}
\definecolor{codepurple}{rgb}{0.58,0,0.82}
\definecolor{backcolour}{rgb}{0.95,0.95,0.92}

\usepackage[normalem]{ulem}
\useunder{\uline}{\ul}{}

\lstdefinestyle{mystyle}{
    backgroundcolor=\color{backcolour},   
    commentstyle=\color{codegreen},
    keywordstyle=\color{magenta},
    numberstyle=\tiny\color{codegray},
    stringstyle=\color{codepurple},
    basicstyle=\ttfamily\footnotesize,
    breakatwhitespace=false,         
    breaklines=true,                 
    captionpos=b,                    
    keepspaces=true,                 
    numbers=left,                    
    numbersep=5pt,                  
    showspaces=false,                
    showstringspaces=false,
    showtabs=false,                  
    tabsize=2
}

\usepackage{booktabs}
\usepackage{multirow}

\hypersetup{
    colorlinks,
    linkcolor={red!50!black},
    citecolor={blue!50!black},
    urlcolor={blue!80!black}
}
 
\theoremstyle{definition}
\newtheorem{definition}{Definition}[section]

\newcommand{\defeq}{\vcentcolon=}
\newcommand{\eqdef}{=\vcentcolon}
\DeclarePairedDelimiterX{\inner}[2]{\langle}{\rangle}{#1, #2}

\DeclarePairedDelimiter\abs{\lvert}{\rvert}%
\DeclarePairedDelimiter\norm{\lVert}{\rVert}%

\title{OnlyDense: Reduced-Order Modeling for Lagrangian simulation}

\author{
  \textbf{Tu T. Do}\thanks{\texttt{s224930346@deakin.edu.au}} \and
  \textbf{Ryan Shannon} \and
  \textbf{Rana Santu} \\[0.5em]
  A$^2$I$^2$, Deakin University, Geelong, Victoria, Australia
}
\date{}

\begin{document}

\maketitle


\begin{abstract}
In science and engineering, Lagrangian simulation methods such as Smooth Particle Hydrodynamics (SPH) or Material Point Method (MPM) are often employed to study the behavior of dynamic systems. However, these methods can be prohibitively computationally expensive, particularly when simulating multi-scale spatial or temporal phenomena, e.g., void growth and coalescence within macro-scale geometries, structural failure of spacecraft components resulting from hypervelocity impact of space debris particles, etc.

%
%
%

In contrast to graph-based methods, where the state of the system is understood as a discrete set of particles,
we propose a learning framework for scalable representation and dynamics modeling of massive particle systems by treating the system state as a function and its evolution as a trajectory in Hilbert space. Rather than representing the state as a discrete set of particles or embedding it in a nonlinear latent manifold, we approximate the state space with a linear subspace spanned by learned neural basis functions. This parameterization enables direct projection to obtain latent coefficients and explicit access to the basis functions, avoiding optimization over a nonlinear latent space.

The resulting representation admits a natural interpretation: latent variables correspond to coefficients in Hilbert space, and basis functions correspond to spatial modes, closely analogous to Proper Orthogonal Decomposition. The framework thus unifies classical projection-based reduced-order modeling with modern deep learning, while remaining invariant to the number of discretization points.

Experiments on large-scale SPH simulations with over one million particles, including dynamic events with extreme deformation and fragmentation, demonstrate that the proposed method achieves accurate reconstruction and dynamics prediction, achieving an R$^2$ score above $0.99$ with as few as $32$ basis functions. A current limitation of the method is its reliance on a single reference configuration, which will be addressed in future work.

%
%

\end{abstract}


\begin{figure}[t]
  \centering
  \subfigure[Monolithic Shield]{\includegraphics[width=.24\textwidth]{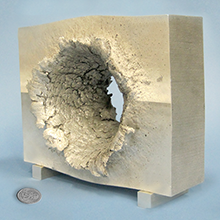}}
  \subfigure[Whipple Shield]{\includegraphics[width=.24\textwidth]{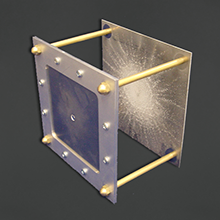}}
  \subfigure[Debris cloud]{\includegraphics[width=.24\textwidth]{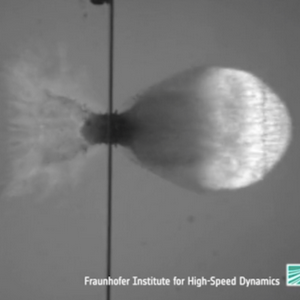}}
  \subfigure[Impact simulation]{\includegraphics[width=.24\textwidth]{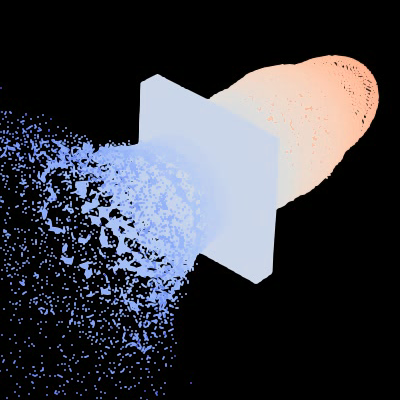}}
  \caption{From left to right: a) a slab of aluminium punctured upon hypervelocity impact; b) the Whipple shield consists of a bumper made of thin aluminium sheet in front of a spacecraft hull; c) imaging of debris cloud formed when a projectile hit the ``meteor bumper''; and d) a simulation of hypervelocity impact with a resolution of approximately 1M particles}
\end{figure}

\section{Introduction}







One of the most prevalent threats to human life and spacecraft in the orbital environment is the collision with micrometeoroids and orbital debris (MMOD). Collisions with MMOD, which have an impacting velocity ranging from a few kilometers per second to tens of kilometers per second, are categorized in the hypervelocity impact regime. These collisions pose a significant threat to the current fleet of satellites and other orbital infrastructure. Therefore, assessing the performance of the spacecraft shield against MMOD is critical. Numerical simulation methods have been the primary tool to study the behavior of the shield under such impact. In the case of hypervelocity impact, where the deformation is large, Lagrangian methods such as Smooth Particle Hydrodynamic (SPH) \cite{gingold1977smoothed} or Material Point Method (MPM) \cite{sulsky_particle_1994} are often employed.
However, there are two limitations associated with numerical simulations, namely the cost of computation and the non-differentiability. The cost of computation scales quadratically with the number of particles, due to the reliance of the estimation of the spatial gradient of each particle on its neighborhood. Therefore, simulations with millions of particles are costly. In the case of an inverse design problem, where we want to find an optimal configuration for a physical system\cite{zhongkai2023}, the nondifferentiable nature of numerical simulation requires a full simulation run at each iteration of configuration. Therefore, the search problem becomes prohibitively expensive.

There is a growing body of literature on machine learning techniques in the study of dynamical systems. However, we identified two gaps in the literature. Firstly, most of the research focuses on Eulerian simulations (i.e., fluid dynamics), due to the grid-based nature of established machine learning architectures such as convolutional neural networks. 
Secondly, the research that focuses on Lagrangian simulations is mostly a graph-based method, precluding application to a large scale of simulations \cite{sanchezgonzalez2018graphnetworkslearnablephysics, kumar2022gnsgeneralizablegraphneural}.

We study reduced-order modeling for Lagrangian simulations under explicit cost–complexity constraints. We ask two questions: how to represent the state of a dynamical system so that computational cost does not scale with the number of discretization points, and how to model the resulting reduced dynamics. Graph-based methods represent the state as a collection of particles, tying both representation and dynamics to the discretization size. In contrast, we represent the state as a function, enabling a compact description whose complexity is controlled independently of resolution.
Therefore, the trajectory of the dynamical system is a series of functions that live in Hilbert space.
Recent works \cite{chen2023cromcontinuousreducedordermodeling, yin2023continuouspdedynamicsforecasting} parameterizes the space of state functions as a nonlinear manifold. In contrast, we restrict the space of state functions to a learned linear subspace spanned by a finite set of basis functions. By learning this set of basis functions, we obtain explicit access to both the basis and their corresponding coefficients, allowing the system state to be encoded via direct projection rather than through specialized encoders \cite{chen2023cromcontinuousreducedordermodeling} or optimization-based auto-decoding \cite{yin2023continuouspdedynamicsforecasting}.
Given this representation, the state of the system is fully characterized by a low-dimensional coefficient vector, and the system dynamics can be learned entirely in this reduced space. We consider both discrete-time dynamics and continuous-time formulations using Neural Ordinary Differential Equations. Similar to prior work, our approach does not require access to governing equations and therefore falls under the category of non-intrusive ROM.

The following sections are organized as follows:
section (\ref{sec:related}) presents related works;
section (\ref{sec:method}) briefly introduces definition and terminologies,then
discusses the formulation, tasks, and limitations of our method;
section (\ref{sec:exp}) details our datasets, metrics, and hardware for the experiments;
and finally section (\ref{sec:conclude}) summarizes our findings and present further discussions.

\section{Related works}
\label{sec:related}




Machine learning techniques have found applications in many areas of continuum mechanics, including fluid and solid mechanics \cite{brunton_2020, Jin_2023}, particularly in employing these techniques to learn dynamics from simulation data.
While extensive work has been conducted on mesh-based simulation data, stemming from the rich history of classical data-driven techniques originating in fluid mechanics \cite{brunton_2020}, such as Proper Orthogonal Decomposition (POD), Dynamic Mode Decomposition (DMD), and the Koopman Operator, comparatively little has been done on mesh-free data, which is more prevalent in solid mechanics due to the nature of large material deformations.

Recent works utilizing graph-based methods \cite{li2019learningparticledynamicsmanipulating,mrowca2018flexibleneuralrepresentationphysics,sanchezgonzalez2018graphnetworkslearnablephysics} or convolution-based methods \cite{fey2018splinecnn,schenck2018spnets,Ummenhofer2020Lagrangian} have been applied to perform simulations or learn dynamics from simulation data.
These methods employed a graph to represent the physical system, where the nodes represent the set of particles, and the edges represent the interaction between the set of particles. The interaction between non-neighboring nodes is modeled via multiple layers of message passing and used as input to predict the future state of each particle.
However, these methods operate at the particle level, providing no pathway for Reduced-Order Modeling (ROM) techniques similar to POD in fluid dynamics. Therefore, direct comparison to graph-based methods at large-scale Lagrangian simulation are not possible due to computational constraints.

Following early methods such as proper orthogonal decomposition (POD), relying on linear methods to identify latent space, machine learning techniques naturally find their application in ROM. Study \cite{fulton_latentspace_2019} employs an autoencoder neural network to project the state of the dynamical system into a lower-dimensional space; then, it performs time-stepping by relying on implicit integration in the latent space. The state of the system is computed by minimizing the elastic potential of the deformed object while constraining the state to be close to the first-order Taylor expansion.

Studies \cite{chen2023modelreductionmaterialpoint, chen2023cromcontinuousreducedordermodeling, yin2023continuouspdedynamicsforecasting} have showcased the capacity of machine learning techniques to reduce the degree of freedom for the mesh-free simulation data.
In contrast to previous studies \cite{fulton_latentspace_2019, lee_model_2019}, which rely on already discretized representations of the system’s state, these methods propose to view the state of the system as a continuous vector field. 
Particularly by parameterizing the state of the system by a class of neural networks called Implicit Neural Representation (INR), which is conditioned by the latent vector, these methods approximate the state space by the product of reference configuration and latent space, where they interpret the latent vector represent the current state of the system.
Studies \cite{chen2023modelreductionmaterialpoint, chen2023cromcontinuousreducedordermodeling} condition the neural network by simply concatenating the latent vector with the coordinate in reference configuration. In contrast, study \cite{yin2023continuouspdedynamicsforecasting} conditions the neural network with a HyperNetwork~\cite{ha2016hypernetworks}.

However, these methods pose a drawback stemming from the lack of access to the representation of INR, which is required to learn the dynamics in the latent space. While studies \cite{chen2023modelreductionmaterialpoint, chen2023cromcontinuousreducedordermodeling} used a specialized architecture to compute the latent vector, 
study \cite{yin2023continuouspdedynamicsforecasting} employed an auto-decoding scheme, which is an optimization problem. The lack of access to the representation of the function in function space led to the complexity in the encoding step of these methods.

Studies \cite{chen2023modelreductionmaterialpoint, chen2023cromcontinuousreducedordermodeling} require access to the governing equation, therefore  fall under the category of intrusive ROM. In contrast, study \cite{yin2023continuouspdedynamicsforecasting} recovers the dynamics from data and therefore falls under the category of non-intrusive ROM. We position our method under the latter category.

\section{Method}
\label{sec:method}


We formulate the problem in a Lagrangian description, where the state of the system is represented as a function defined on a reference configuration. Consequently, the system trajectory can be viewed as a sequence of functions. We assume that these functions lie in a Hilbert space equipped with an inner product, which enables projection-based representations. Our goal is to learn a compact set of basis functions spanning this space, such that the system state can be efficiently represented and evolved in a low-dimensional coefficient space. See appendix \ref{apdx:continuum_mech} for details about Lagrangian description and reference configuration, and appendix \ref{apdx:inner_product} 

\subsection{Formulation}

Let $\mathfrak{B}$ be some physical body that occupies reference configuration $\Omega_0$ at $t=0$. After some time $t$, the body underwent deformation and occupies the current configuration $\Omega_t$. Let $x, u$ be the Lagrangian description of position and other quantities of interest in a physical system (such as velocity, temperature, pressures)

\begin{equation}
  \begin{aligned}
  x : \Omega_0 \times [0, T] \rightarrow \Omega_t \subset \mathbb{R}^3    &\quad x = x(X, t)\\
  u : \Omega_0 \times [0, T] \rightarrow U \subset \mathbb{R}^d &\quad u = u(X, t)
  \end{aligned}
\end{equation}

where $X \in \Omega_0$ is the Lagrangian coordinate of the system. Then the state of the dynamical system can be characterized by a function $s: \Omega_0\times[0, T] \rightarrow \Omega_t \times U \subset \mathbb{R}^{d + 3}$

\begin{equation}
  s_t \defeq s(X, t) = \begin{bmatrix}x(X, t) & u(X, t)\end{bmatrix}
\end{equation}

The series of functions $\{s_t\}_{t\in[0, T]}$ represents the trajectory of the dynamical system in an infinite-dimensional function space $\mathcal{S}$. We approximate $\mathcal{S}$ by a finite dimensional function space $\mathcal{S}^K = \text{span}\{\phi_1, \phi_2, \cdots \phi_K\}$, where $\{\phi_1, \phi_2, \cdots, \phi_K\}$ is the set of basis functions. Therefore, any function in this space can be written as,

\begin{equation}
  s_t = \sum_{k=1}^K{c_k(t) \phi_k(X)}
\end{equation}

Given the set of basis functions, the set of corresponding coefficients uniquely represents the state of the dynamical system. Our methods consist of two tasks:

\begin{enumerate}
  \item {\it Learning the set of basis functions}: Approximate the set of basis functions with a set of corresponding neural networks $\{\phi_k(\cdot, \theta_k)\}_{k=1}^K$. Where $\Theta =\{\theta_k\}_{k=1}^K$ is the set of basis functions' parameters, we employed the method of Function Encoder \cite{ingebrand2025functionencodersprincipledapproach} to learn the set of parameters $\theta$. And the choice for the neural network will be Coordinate MLP (i.e., SIREN, FourierFeatureNet).

  \item {\it Learning the dynamical function}: Given state $f(t)$ and the set of basis functions $\{\phi_k\}_{k=1}^K$, we can determine the corresponding coefficient $c(t) = \begin{bmatrix}c_1(t) & c_2(t) & \cdots & c_K(t)\end{bmatrix}$ by either taking the inner product of $s$ with corresponding basis or the method of least-square. Having computed $\mathbf{c}$, we wish to learn the dynamic,

  \begin{equation}
    \begin{aligned}
      \text{In case of continuous time}: \quad & \dot{\mathbf{c}}(t) = f(\mathbf{c}, t; \Psi)\\
      \text{In case of discrete time}  : \quad & \mathbf{c}_{t+1} = f(\mathbf{c}_t; \Psi)
    \end{aligned}
  \end{equation}
  Where $\Psi$ is the parameter of the dynamical function.

\end{enumerate}

\subsection{Learning the set of basis functions}

Given the trajectory of the dynamical system $\{s_t\}_{t\in[0, T]}$, this trajectory lives in a space that is spanned by the set of basis functions $\{\phi_k\}_{k=1}^K$. We employ Function Encoder \cite{ingebrand2025functionencodersprincipledapproach} to approximate the set of basis functions by a corresponding set of neural networks $\{\phi_k(\dot, \theta_k)\}_{k=1}^K$. Where $\Theta = \{\theta_k\}_{k=1}^K$ is the set of parameters.
For each basis, we estimate the coefficient of the corresponding basis by either projecting the state onto each basis by taking the inner product,

\begin{equation}
  \label{eq:coef_inner}
  \begin{aligned}
    \hat{c}_k(t) & = \inner{s_t}{\phi_k} = \int_{X\in\Omega_0}{
      s(X, t) \phi_k(X; \theta_k) dX
    }\\
    &\approx \frac{1}{N}\sum_{i=1}^N{
      s(X^{(i)}, t)\phi_k(X^{(i)};\theta_k)
    } = \frac{1}{N}
      \sum_{i=1}^N{s^{(i)}_t \phi_k(X^{(i)};\theta_k)}
  \end{aligned}
\end{equation}

Or by solving a system of linear equations using the ordinary least squares method, where we organize our data into matrix form

\begin{equation}
  \label{eq:coef_lsq}
  S(t) = \begin{bmatrix}
    s^{(1)}_t\\
    s^{(2)}_t\\
    \vdots\\
    s^{(N)}_t
  \end{bmatrix} = 
  \underbrace{
    \begin{bmatrix}
      \phi_1(X^{(1)}) & \cdots & \phi_K(X^{(1)})\\
      \phi_1(X^{(2)}) & \cdots & \phi_K(X^{(2)})\\
      \vdots          & \ddots & \vdots\\
      \phi_1(X^{(N)}) & \cdots & \phi_K(X^{(N)})\\
    \end{bmatrix}}_{\eqdef \Phi \in \mathbb{R}^{N \times K}}
  \underbrace{
    \begin{bmatrix}
      c_1(t)\\
      c_2(t)\\
      \vdots\\
      c_K(t)
    \end{bmatrix}}_{\eqdef \mathbf{c}(t) \in \mathbb{R}^K}
\end{equation}

Then $\mathbf{c}(t)$ can be obtained by the normal equation,

\begin{equation}
  \label{eq:coef_lsq}
  \mathbf{c}(t) = (\Phi^\top\Phi)^{-1}\Phi^\top S(t)
\end{equation}

Where, $\{(X^{(i)}, s^{(i)}_t)\}_{i=1}^N$ is the set of discretization points of the system state $s(X, t):\forall X\in\Omega_0$. Then we can recover the state of the system from the set of basis functions and their corresponding coefficients,

\begin{equation}
  \label{eq: reconstruct}
  \hat{s}(X, t) = \sum_{k=1}^K{\hat{c}_k(t)\phi_k(X;\theta_k)}
\end{equation}

We obtain the set of parameters by minimizing the distance between the reconstructed vector field and its groundtruth.

\begin{equation}
  \Theta = \text{argmin}_{\Theta} \norm{s(X, t) - \hat{s}(X, t)}
\end{equation}

The cost of evaluating both equation \ref{eq:coef_inner} and \ref{eq:coef_lsq}, which scales linearly with the number of particles $\mathcal{O}(N \times K)$ and $\mathcal{O}(N\times K^2 + K^3$ respectively, can be expensive when we discretize the computation domain into million of particles. However, we can just sample $M$ out of $N$ particles to evaluate both equations, where $N \gg M \ge K$ because the computation of the coefficients can be viewed as either taking the inner product or solving the ordinary least squares problem.
For the inner product projection, the integral can be approximated using Monte Carlo integration by sampling $M < N$ points from $\Omega_0$. In contrast, when solving for the coefficients via a system of linear equations, only $K \ll N$ unknowns are involved. As a result, the system is overparameterized, and sampling $M \ge K$ points ensures that the system is well defined. A similar argument appears in \cite{chen2023cromcontinuousreducedordermodeling}.
The detailed algorithm of the Function Encoder could be found in appendix \ref{apdx:fe_algo}.

\subsection{Learning the dynamic function}

As mentioned in the formulation, we can view our dynamical system as either discrete time or continuous time. 

\paragraph{\emph Discrete time}{
  In the case of a discrete-time dynamical system, we want to learn the dynamic function $f$ that pushes the state of the system one time step. We parameterize the dynamic function $f$ by a fully connected neural network $f(\cdot;\Psi)$. Given the state of the system $c_j$ and $c_{j+1}$ at time step $t_j$ and $t_{j+1}$ respectively, we predict the next state by,
  \begin{equation}
    \label{eq:discrete_dynamic}
    \hat{\mathbf{c}}_{j+1}=f(\mathbf{c}_j;\Psi)
  \end{equation}
  Then we obtain the parameter $\Psi$ by minimizing,
  \begin{equation*}
    \begin{aligned}
      \Psi &= \argmin_\Psi \norm {\mathbf{c}_{j+1} - \hat{\mathbf{c}}_{j+1}}\\
      &= \argmin_\Psi \norm{\mathbf{c}_{j+1} - f(\mathbf{c}_j;\Psi)}
    \end{aligned}
  \end{equation*}
  , where
  \begin{equation*}
    \mathbf{c}_j=\begin{bmatrix}
      \inner{s_j}{\phi_1}\\
      \cdots\\
      \inner{s_j}{\phi_K}\\
    \end{bmatrix}\quad
    \mathbf{c}_{j+1}=\begin{bmatrix}
      \inner{s_{j+1}}{\phi_1}\\
      \cdots\\
      \inner{s_{j+1}}{\phi_K}\\
    \end{bmatrix}
  \end{equation*}
}

\paragraph{Continuous time}{ When we view the system as a continuous time dynamic system, we wish to approximate the time derivative of the state. We approximate the time derivative of the state by a neural network $f(\cdot, \Psi)$,

  \begin{equation*}
    \dot{c}(t) = f(c, t; \Psi)
  \end{equation*}

  Given the state of the system $\mathbf{c}_i$ and $\mathbf{c}_j$ at $t_i$ and $t_j$, we predict the state at $t_j$ by,

  \begin{equation}
    \label{eq:continuous_dynamic}
    \hat{\mathbf{c}}_{j} = \mathbf{c}_i + \int_{t_i}^{t_j}{
        f(\dot{c}, t;\Psi) dt
      }
  \end{equation}
  Then we obtain the parameter $\Psi$ by minimizing,
  \begin{equation*}
    \begin{aligned}
    \Psi &= \argmin_\Psi \norm {\mathbf{c}_j - \hat{\mathbf{c}}_j}\\
    &= \argmin_\Psi \bigg\lVert 
      \mathbf{c}_j - \big(\mathbf{c}_i + \int_{t_i}^{t_j}{
        f(\dot{c}, t;\Psi) dt
      }\big)
    \bigg\rVert 
    \end{aligned}
  \end{equation*}

  , where

  \begin{equation*}
    \mathbf{c}_i=\begin{bmatrix}
      \inner{s_i}{\phi_1}\\
      \cdots\\
      \inner{s_i}{\phi_K}\\
    \end{bmatrix}\quad
    \mathbf{c}_j=\begin{bmatrix}
      \inner{s_j}{\phi_1}\\
      \cdots\\
      \inner{s_j}{\phi_K}\\
    \end{bmatrix}
  \end{equation*}

  This is a standard formulation for Neural Ordinary Differential Equations \cite{chen2019}. We obtain the parameters $\Psi$ by the adjoint state method.
}

\subsection{Sharing basis between position and velocity fields}

Assume that the position vector field can be described in terms of a set of basis functions $\{\phi_k\}_{k=1}^K$, where $\phi_k: \Omega_0\rightarrow\mathbb{R}^3$. Then the position function can be written as,

\begin{equation}
  x_t \defeq x(X, t) = \sum_{k=1}^K{
    c_k(t)\phi_k(X)
  }
\end{equation}

Taking the derivative of both sides of the equation with respect to time gives us the velocity vector field,

\begin{equation}
  v_t \defeq \dot{x}(X, t) = \sum_{k=1}^K{
    \dot{c}_k(t)\phi_k(X)
  }
\end{equation}

Therefore, the velocity vector field has to lie within the space spanned by the set of basis functions $\{\phi_k\}_{k=1}^K$.

\subsection{Limitation}
Since the definition of the function space where the trajectory of the system lives, and the definition of the inner product defined in said space, relied on the shared domain of all functions within the space.
While this method will work in the Eulerian setting where the computation domain is fixed,
a limitation in the Lagrangian setting is that it can not generalize to more than one reference configuration and we are addressing this in future work. 

%

\section{Experiments}


\label{sec:exp}

\subsection{Datasets}

We organize our dataset into two groups, namely the Eulerian and Lagrangian datasets. The first group consists of the Navier-Stokes equation and the Wave equation, which have been included in previous methods that share the same view that the state of the system is a function. The second group consists of simulations in Lagrangian settings
We include the first group to compare our method to existing methods, and the second group to test the scalability of our approach. Furthermore, we include the first group of datasets to test whether our method also works in an Eulerian setting despite being initially formulated in the Lagrangian setting.

\paragraph{\emph 2D Wave equation and 2D Navier Stokes equation}{
  We included two simulations in an Eulerian setting, namely the \emph{2D Wave equation} and the \emph{2D Navier Stokes equation}. The 2D Wave equation is a second-order PDE $\frac{\partial^2u}{\partial t} = c^2\Delta u$, where $u$ is the displacement with respect to the rest position and $c$ is the traveling speed. The 2D Navier Stokes equation corresponds to an incompressible fluid dynamics $\frac{dv}{dt}=-u\nabla v+\nu\Delta v + f; v = \nabla \times u; \nabla u = 0$, where $u$ is the velocity field, $v$ is the vorticity field, $\nu$ is the viscosity and $f$ is a constant forcing term. Both equations were solved on a $64\times64$ grid with temporal resolutions of $\Delta t=0.25$ second and $1.0$ second for the Wave equation and Navier-Stokes equation, respectively.
}

\paragraph{\emph Exploding }{Exploding plate is a simulation of approximately $22$ thousand particles over a trajectory of 500 time steps, with an initial geometry of a box. The initial height, depth, and width of the plate are $(1.0, 1.0, 0.2) \times 10^{-3}$ meters. The spatial and temporal resolutions are $\Delta x=2\times10^{-5}$ and $\Delta t = 7\times10^{-7}$ second. The initial vector for each particle is given by,

  \begin{equation*}
    \begin{aligned}
        \begin{bmatrix}
            v_x\\
            v_y\\
            v_z
        \end{bmatrix} = 600\times\begin{bmatrix}
          \sin(2000 \pi x)\\
          \sin(2000 \pi y)\\
          0
        \end{bmatrix}\\
    \end{aligned}
  \end{equation*}
}

\paragraph{\emph Mott ring}{Mott ring is a simulation of a hollow cylinder discretized into about $3.4$ thousand particles. The initial inner radius, outer radius, and thickness of the cylinder are $(6.0, 6.75, .75)\times10^{-2}$ meters. The spatial and temporal resolutions are $\Delta x=1.875\times10^{-3}$ and $\Delta t=1.4\times10^{-7}$. The initial velocity of each particle is given by,

  \begin{equation*}
    \begin{bmatrix}
      v_x\\
      v_y\\
      v_z
    \end{bmatrix} = 600\times\begin{bmatrix}
      \frac{x}{\sqrt{x^2+y^2}}\\
      \frac{y}{\sqrt{x^2+y^2}}\\
      0
    \end{bmatrix}
  \end{equation*}
  }

\paragraph{\emph Projectile}{
  Projectile is a simulation of an aluminum projectile hitting a steel plate in a hypervelocity impact regime over $400$ time steps. The simulation discretizes the projectile and the plate into approximately $1.1$ million particles. The radius of the projectile is $1.5\times10^{-3}$ meter; the height, width, and thickness of the plate are $(2.0, 2.0, 0.3) \times 10^{-2}$ meter, respectively. The spatial and temporal resolution are $\Delta x=10^{-4}$ meter and $\Delta t=2.5\times10^{-9}$ second. In the initial condition, the projectile's velocity is approximately $7.3 km/s$ when it collides with the static plate at some small oblique angle. All simulations are run in ABSTRAO \cite{Coll2019} and the parameters are summarized in the following table,
}

\begin{table}[H]
  \centering
  \begin{tabular}{lccccc}
    \toprule
    Dataset         & N       & L           & M         & $\Delta x$         & $\Delta t$         \\ \midrule
    Mott ring       & $3.4K$  & $961$/$320$ & $1024$    & $1.9\times10^{-3}$ & $1.4\times10^{-7}$ \\
    Exploding       & $22K$   & $375$/$125$ & $2048$    & $2.0\times10^{-5}$ & $7.0\times10^{-7}$ \\
    Projectile      & $1.1M$  & $300$/$100$ & $2048$    & $1.0\times10^{-4}$ & $2.5\times10^{-9}$ \\ \bottomrule
  \end{tabular}
  \caption{Number of particles per time-step (N), number of time-steps per trajectory (L), and number of particles sampled per gradient step (S)}
\label{tab:dataset}
\end{table}

\subsection{Metrics} Given function $u, v: \Omega\rightarrow \mathbb{R}$, we defined the following error measurements. In the general case where $u,v$ are vector-valued functions, the measurements are averaged across the dimension of output space.

\paragraph{MSE}{Mean-squared error between two vector fields is given by,

\begin{equation*}
  \text{MSE}(u, v) \defeq \int_{x\in\Omega}{
    (u(x) - v(x))^2dx
  } \approx \frac{1}{N}\sum_{i=1}^N (u_i - v_i)^2
\end{equation*}

}

\paragraph{MAE}{Mean-absolute error between two vector fields is given by,

\begin{equation*}
  \text{MAE}(u, v) \defeq \int_{x\in\Omega}{
    \abs{u(x) - v(x)} dx
  } \approx \frac{1}{N}\sum_{i=1}^N \abs{u_i - v_i}
\end{equation*}

}

\paragraph{R-square}{ Given ground-truth $u$ and its prediction $v$, where $u, v: \Omega \rightarrow \mathbb{R}$, the coefficient of determination, is given by 
  \begin{equation}
    \text{R}^2(u, v) \defeq 1 - \frac{\text{MSE}(u, v)}{\text{MSE}(u, \bar{u})}
  \end{equation}

  Where $\bar{u} = \int_{x\in\Omega}{u(x) dx}$. When the co-domain is high-dimensional, the score is averaged across dimensions, weighted by its respective variance.
}

\subsection{Results}

\subsubsection{Comparison with baseline}

We evaluate OnlyDense on two datasets, the \emph{2D Navier--Stokes equation} and the \emph{2D Wave equation}, both solved on a fixed $64 \times 64$ grid. Performance is reported under both \emph{in-time (In-T)} and \emph{out-of-time (Out-T)} settings, where In-T evaluates predictions within the training time horizon, and Out-T evaluates temporal extrapolation beyond the training window, a regime in which error accumulation and stability become critical.

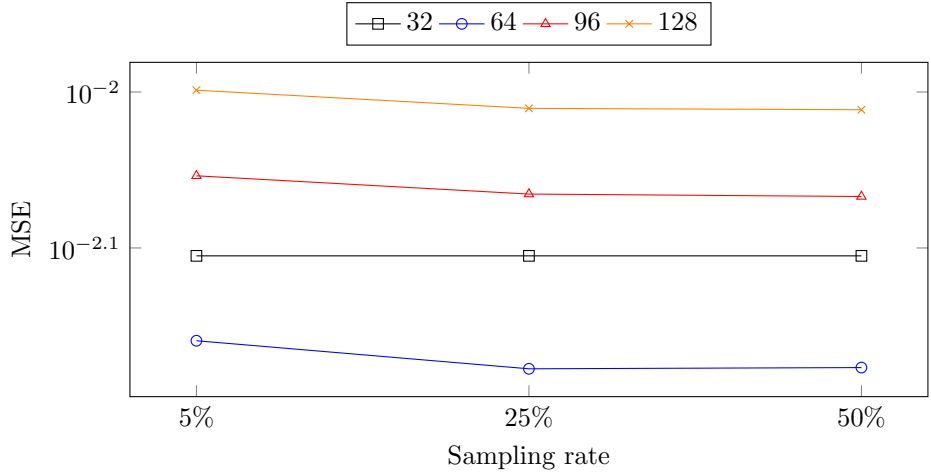
\begin{figure}[h]
\centering
\begin{tikzpicture}
\begin{axis}[
    legend style={at={(0.5,1.05)}, anchor=south},
    legend columns=-1, 
    width=\columnwidth,
    xlabel={Sampling rate},
    ylabel={MSE},
    ymode=log,
    height=6cm,
    symbolic x coords={$5\%$, $25\%$, $50\%$},
    xtick=data,
]
\addplot [
  color=black,
  mark=square
]coordinates {
($5\%$,   0.007851)
($25\%$,  0.007851)
($50\%$,  0.007851)
};
\addlegendentry{$32$};

\addplot [
  color=blue,
  mark=o
]coordinates {
($5\%$,   0.006925)
($25\%$,  0.006644)
($50\%$,  0.006656)
};
\addlegendentry{$64$};

\addplot [
  color=red,
  mark=triangle
]coordinates {
($5\%$,   0.008834)
($25\%$,  0.008599)
($50\%$,  0.008568)
};
\addlegendentry{$96$};
 
\addplot [
  color=orange,
  mark=x
]coordinates {
($5\%$,   0.010024)
($25\%$,  0.009759)
($50\%$,  0.009739)
};
\addlegendentry{$128$};
\end{axis}
\end{tikzpicture}
\caption{MSE Out-T across different spatial sampling rate and number of basis function on 2D Navier Stokes equation.}
\label{fig:mse_navierstokes}
\end{figure}

\begin{figure}[h]
\centering
\begin{tikzpicture}
\begin{axis}[
    legend style={at={(0.5,1.05)}, anchor=south},
    legend columns=-1, 
    width=\columnwidth,
    xlabel={Sampling rate},
    ylabel={MSE},
    ymode=log,
    height=6cm,
    symbolic x coords={$5\%$, $25\%$, $50\%$},
    xtick=data,
]
\addplot [
  color=black,
  mark=square
]coordinates {
($5\%$,   0.014899)
($25\%$,  0.014504)
($50\%$,  0.014296)
};
\addlegendentry{$32$};

\addplot [
  color=blue,
  mark=o
]coordinates {
($5\%$,   0.007449)
($25\%$,  0.007127)
($50\%$,  0.007100)
};
\addlegendentry{$64$};

\addplot [
  color=red,
  mark=triangle
]coordinates {
($5\%$,   0.006645)
($25\%$,  0.006672)
($50\%$,  0.006530)
};
\addlegendentry{$96$};
 
\addplot [
  color=orange,
  mark=x
]coordinates {
($5\%$,   0.005772)
($25\%$,  0.005657)
($50\%$,  0.005693)
};
\addlegendentry{$128$};
\end{axis}
\end{tikzpicture}
\caption{MSE Out-T across different spatial sampling rate and number of basis function on 2D Wave equation.}
\label{fig:mse_wave}
\end{figure}
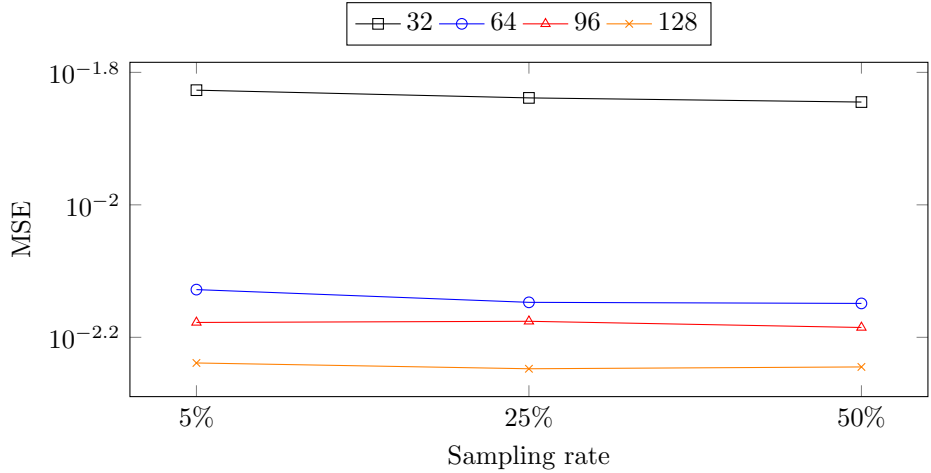

To assess robustness to partial observations, we vary the \emph{spatial sampling rate} during training ($5\%$, $25\%$, and $50\%$), following DINo’s emphasis on generalization under sparse spatial measurements \cite{yin2023continuouspdedynamicsforecasting}. Across both benchmarks, OnlyDense exhibits stable performance across sampling rates, with relatively small degradation as the number of observed points decreases, in both In-T and Out-T evaluations.
While some baseline methods achieve lower absolute error at higher sampling rates, their performance degrades more sharply under sparse sampling. In contrast, OnlyDense’s projection-based encoding into a \emph{learned linear latent space} yields a reduced representation that remains well conditioned even with limited observations as demonstrated in figure \ref{fig:mse_navierstokes} and \ref{fig:mse_wave}. These results highlight OnlyDense as a scalable and sampling-efficient alternative for spatiotemporal dynamics modeling, particularly suited to large systems or regimes where dense observations are unavailable.

Numbers reported in table \ref{tb:baseline_combined} are results where we approximate the state space with $128$ basis functions, baseline results are taken from \cite{yin2023continuouspdedynamicsforecasting}. For complete results, please see table \ref{tb:baseline_full} in appendix \ref{apdx:more_result}.

\begin{table}[H]
\resizebox{\columnwidth}{!}{
\begin{tabular}{clllll}
\hline
\multirow{2}{*}{\begin{tabular}[c]{@{}c@{}}Sample\\ size\end{tabular}} 
& \multirow{2}{*}{Model} 
& \multicolumn{2}{c}{Train} 
& \multicolumn{2}{c}{Test} \\ \cline{3-6}
& & \multicolumn{1}{c}{In-t} 
& \multicolumn{1}{c}{Out-t} 
& \multicolumn{1}{c}{In-t} 
& \multicolumn{1}{c}{Out-t} \\ 
\hline

\multicolumn{6}{c}{\textbf{2D Navier--Stokes}} \\ \hline

\multirow{5}{*}{5\%}  & I-MP-PDE  & $\num{8.154E-03}$ & $\num{8.166E-03}$ & $\num{7.926E-03}$ & $\num{8.225E-03}$ \\
                      & DeepONet  & $\num{3.330E-03}$ & $\num{7.370E-03}$ & $\num{1.345E-02}$ & $\num{1.408E-02}$ \\
                      & SIREN     & $\num{8.741E-03}$ & $\num{1.767E-01}$ & $\num{4.303E-02}$ & $\num{2.126E-01}$ \\
                      & DINo      & $\num{1.029E-03}$ & $\num{1.655E-03}$ & $\num{1.326E-03}$ & $\num{1.813E-03}$ \\
                      & OnlyDense & $\num{3.578E-03}$ & $\num{8.494E-03}$ & $\num{4.592E-03}$ & $\num{1.002E-02}$ \\ \hline

\multirow{5}{*}{25\%} & I-MP-PDE  & $\num{3.135E-04}$ & $\num{7.245E-04}$ & $\num{3.476E-04}$ & $\num{7.658E-04}$ \\
                      & DeepONet  & $\num{9.016E-04}$ & $\num{5.936E-03}$ & $\num{9.376E-03}$ & $\num{1.328E-02}$ \\
                      & SIREN     & $\num{5.180E-03}$ & $\num{2.175E-01}$ & $\num{2.436E-01}$ & $\num{3.861E-01}$ \\
                      & DINo      & $\num{1.020E-04}$ & $\num{4.504E-04}$ & $\num{2.646E-04}$ & $\num{5.951E-04}$ \\
                      & OnlyDense & $\num{3.258E-03}$ & $\num{8.142E-03}$ & $\num{4.296E-03}$ & $\num{9.759E-03}$ \\ \hline

\multirow{5}{*}{50\%} & I-MP-PDE  & $\num{1.170E-04}$ & $\num{5.013E-04}$ & $\num{1.611E-04}$ & $\num{6.088E-04}$ \\
                      & DeepONet  & $\num{6.541E-04}$ & $\num{4.311E-03}$ & $\num{5.720E-03}$ & $\num{1.091E-02}$ \\
                      & SIREN     & $\num{4.995E-03}$ & $\num{6.841E-01}$ & $\num{1.603E-01}$ & $\num{6.876E-01}$ \\
                      & DINo      & $\num{8.677E-05}$ & $\num{2.962E-04}$ & $\num{2.062E-04}$ & $\num{4.348E-04}$ \\
                      & OnlyDense & $\num{3.256E-03}$ & $\num{8.193E-03}$ & $\num{4.280E-03}$ & $\num{9.739E-03}$ \\ \hline

\multicolumn{6}{c}{\textbf{2D Wave}} \\ \hline

\multirow{5}{*}{5\%}  & I-MP-PDE  & $\num{7.055E-04}$ & $\num{7.097E-04}$ & $\num{1.138E-03}$ & $\num{1.116E-03}$ \\
                      & DeepONet  & $\num{8.331E-04}$ & $\num{9.295E-03}$ & $\num{1.692E-02}$ & $\num{3.256E-02}$ \\
                      & SIREN     & $\num{2.738E-03}$ & $\num{1.818E-02}$ & $\num{3.339E-02}$ & $\num{6.964E-02}$ \\
                      & DINo      & $\num{4.088E-05}$ & $\num{4.121E-05}$ & $\num{6.415E-05}$ & $\num{7.392E-05}$ \\
                      & OnlyDense & $\num{9.440E-05}$ & $\num{1.016E-03}$ & $\num{8.906E-04}$ & $\num{5.772E-03}$ \\ \hline

\multirow{5}{*}{25\%} & I-MP-PDE  & $\num{3.293E-05}$ & $\num{1.108E-04}$ & $\num{5.142E-05}$ & $\num{1.545E-04}$ \\
                      & DeepONet  & $\num{5.722E-04}$ & $\num{1.061E-02}$ & $\num{1.757E-02}$ & $\num{3.221E-02}$ \\
                      & SIREN     & $\num{8.995E-04}$ & $\num{1.292E-02}$ & $\num{1.783E-02}$ & $\num{5.143E-02}$ \\
                      & DINo      & $\num{3.949E-06}$ & $\num{4.436E-06}$ & $\num{1.089E-05}$ & $\num{1.174E-05}$ \\
                      & OnlyDense & $\num{8.877E-05}$ & $\num{1.001E-03}$ & $\num{8.650E-04}$ & $\num{5.657E-03}$ \\ \hline

\multirow{5}{*}{50\%} & I-MP-PDE  & $\num{6.021E-07}$ & $\num{3.166E-05}$ & $\num{1.646E-06}$ & $\num{6.710E-05}$ \\
                      & DeepONet  & $\num{7.715E-04}$ & $\num{1.186E-02}$ & $\num{1.665E-02}$ & $\num{3.385E-02}$ \\
                      & SIREN     & $\num{5.246E-04}$ & $\num{1.486E-02}$ & $\num{3.100E-02}$ & $\num{8.265E-02}$ \\
                      & DINo      & $\num{3.380E-06}$ & $\num{3.751E-06}$ & $\num{9.251E-06}$ & $\num{9.710E-06}$ \\
                      & OnlyDense & $\num{8.940E-05}$ & $\num{1.001E-03}$ & $\num{8.724E-04}$ & $\num{5.693E-03}$ \\ \hline
\end{tabular}
}
\caption{Comparison of OnlyDense with other methods on the 2D Navier--Stokes and 2D Wave equations.}
\label{tb:baseline_combined}
\end{table}


\subsubsection{Result on ABSTRAO's simulations}

To reiterate our research questions, we want to examine
\emph{(i) basis learning} whether Function Encoder represents the state of the dynamical system accurately, and
\emph{(ii) dynamic learning} whether we can learn the dynamics from data given the representation. In order to answer these questions, we evaluated the reconstruction loss for the current state and future state for the \emph{basis learning task} and \emph{dynamic learning task}, respectively. 
For the evaluation of both tasks, the state of the system is determined by computing the coefficients with a subset of particles, and then the prediction is evaluated over the entire set of particles.
The sampling in the computation of the coefficient step introduces randomness into our evaluation, therefore we evaluate all configurations across $4$ different random seeds.

\paragraph{Learning set of basis functions}{
At each time step, the coefficient $\mathbf{c}$ is computed using either equation \ref{eq:coef_inner} or \ref{eq:coef_lsq}.
Then the state function is recovered from the set of basis functions and their corresponding coefficients by equation \ref{eq: reconstruct}.
The reconstruction loss is then evaluated by comparing the evaluation of the state function and the ground truth over the entire set of points.

  \begin{table}[h]
  \begin{tabular}{
      >{\arraybackslash}p{1.25cm}
      >{\centering\arraybackslash}p{1cm}
      c
      c
    }
    \toprule
    &  & Train R$^2$ & Test R$^2$ \\
    Dataset & Basis ($K$) &  &  \\
    \midrule
    \multirow[t]{11}{*}{Exploding} & 2 & $\num{98.3416} \quad\hfill ( \pm \num{0.2419} )$ & $\num{98.6247} \quad\hfill ( \pm \num{0.0188} )$ \\
    & 4 & $\num{99.2290} \quad\hfill ( \pm \num{0.1218} )$ & $\num{99.2763} \quad\hfill ( \pm \num{0.0358} )$ \\
    & 8 & $\num{99.8650} \quad\hfill ( \pm \num{0.0255} )$ & $\num{99.8882} \quad\hfill ( \pm \num{0.0043} )$ \\
    & 16 & $\num{99.9831} \quad\hfill ( \pm \num{0.0098} )$ & $\num{99.9950} \quad\hfill ( \pm \num{0.0020} )$ \\
    & 32 & $\num{99.9983} \quad\hfill ( \pm \num{0.0006} )$ & $\num{99.9991} \quad\hfill ( \pm \num{0.0001} )$ \\
    & 64 & $\num{99.9991} \quad\hfill ( \pm \num{0.0003} )$ & $\num{99.9992} \quad\hfill ( \pm \num{0.0001} )$ \\
    & 96 & $\num{99.9993} \quad\hfill ( \pm \num{0.0002} )$ & $\num{99.9995} \quad\hfill ( \pm \num{0.0001} )$ \\
    \cline{1-4}
    \multirow[t]{11}{*}{Mott ring} & 2 & $\num{99.7917} \quad\hfill ( \pm \num{0.0994} )$ & $\num{99.9107} \quad\hfill ( \pm \num{0.0104} )$ \\
    & 4 & $\num{99.9446} \quad\hfill ( \pm \num{0.0281} )$ & $\num{99.9678} \quad\hfill ( \pm \num{0.0123} )$ \\
    & 8 & $\num{99.9667} \quad\hfill ( \pm \num{0.0124} )$ & $\num{99.9899} \quad\hfill ( \pm \num{0.0025} )$ \\
    & 16 & $\num{99.9886} \quad\hfill ( \pm \num{0.0120} )$ & $\num{99.9973} \quad\hfill ( \pm \num{0.0002} )$ \\
    & 32 & $\num{99.9878} \quad\hfill ( \pm \num{0.0066} )$ & $\num{99.9985} \quad\hfill ( \pm \num{0.0002} )$ \\
    & 64 & $\num{99.9968} \quad\hfill ( \pm \num{0.0007} )$ & $\num{99.9986} \quad\hfill ( \pm \num{0.0011} )$ \\
    & 96 & $\num{99.9962} \quad\hfill ( \pm \num{0.0034} )$ & $\num{99.9990} \quad\hfill ( \pm \num{0.0006} )$ \\
    \cline{1-4}
    \multirow[t]{11}{*}{Projectile} & 2 & $\num{88.7186} \quad\hfill ( \pm \num{5.1813} )$ & $\num{94.8356} \quad\hfill ( \pm \num{1.7446} )$ \\
    & 4 & $\num{97.6520} \quad\hfill ( \pm \num{0.3678} )$ & $\num{97.9315} \quad\hfill ( \pm \num{0.1017} )$ \\
    & 8 & $\num{98.5252} \quad\hfill ( \pm \num{0.0934} )$ & $\num{98.6370} \quad\hfill ( \pm \num{0.0427} )$ \\
    & 16 & $\num{98.6602} \quad\hfill ( \pm \num{0.2050} )$ & $\num{98.6682} \quad\hfill ( \pm \num{0.0859} )$ \\
    & 32 & $\num{99.0982} \quad\hfill ( \pm \num{0.0516} )$ & $\num{99.0388} \quad\hfill ( \pm \num{0.0930} )$ \\
    & 64 & $\num{98.8312} \quad\hfill ( \pm \num{0.4004} )$ & $\num{99.1914} \quad\hfill ( \pm \num{0.1772} )$ \\
    & 96 & $\num{98.8706} \quad\hfill ( \pm \num{0.2491} )$ & $\num{99.1768} \quad\hfill ( \pm \num{0.1014} )$ \\
    \cline{1-4}
    \bottomrule
  \end{tabular}
  \caption{Average and standard deviation of R$^2$ ($\mu \pm \sigma$) across different datasets and number of basis functions on basis learning task.}
  \label{tb:basis_r2}
\end{table}

  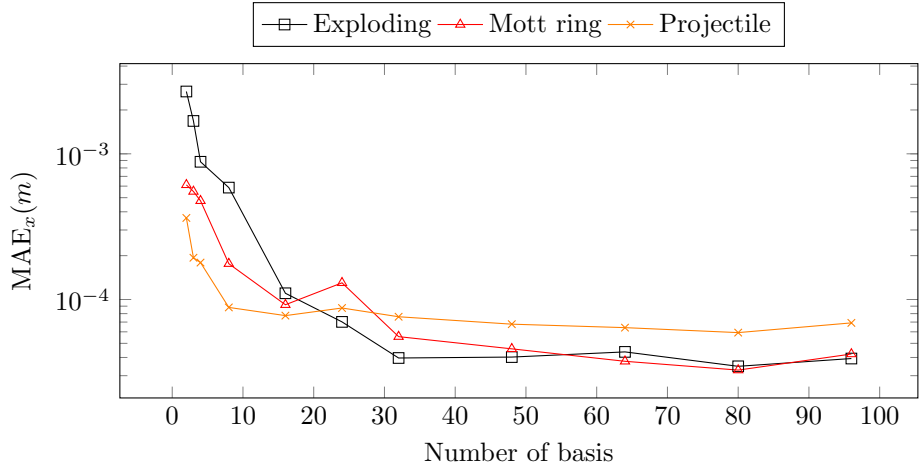
\begin{figure}[h]
\centering
\begin{tikzpicture}
\begin{axis}[
    legend style={at={(0.5,1.05)}, anchor=south},
    legend columns=-1, 
    width=\columnwidth,
    xlabel={Number of basis},
    ylabel={MAE$_x (m)$},
    ymode=log,
    height=6cm
]
\addplot [
  color=black,
  mark=square
]coordinates {
  (2, 2.6777e-03)
    (3, 1.6809e-03)
    (4, 8.8162e-04)
    (8, 5.8627e-04)
    (16, 1.1035e-04)
    (24, 7.0200e-05)
    (32, 3.9674e-05)
    (48, 4.0258e-05)
    (64, 4.3687e-05)
    (80, 3.4788e-05)
    (96, 3.9341e-05)
};
\addlegendentry{Exploding};


\addplot [
  color=red,
  mark=triangle
]coordinates {
(2, 6.1222e-04)
(3, 5.5120e-04)
(4, 4.7579e-04)
(8, 1.7610e-04)
(16, 9.2040e-05)
(24, 1.3016e-04)
(32, 5.5601e-05)
(48, 4.5848e-05)
(64, 3.7661e-05)
(80, 3.2771e-05)
(96, 4.2244e-05)
};
\addlegendentry{Mott ring};

\addplot [
  color=orange,
  mark=x
]coordinates {
(2, 3.6298e-04)
(3, 1.9348e-04)
(4, 1.7926e-04)
(8, 8.8244e-05)
(16, 7.7519e-05)
(24, 8.7427e-05)
(32, 7.6153e-05)
(48, 6.7719e-05)
(64, 6.4103e-05)
(80, 5.9260e-05)
(96, 6.9102e-05)
};
\addlegendentry{Projectile};
\end{axis}
\end{tikzpicture}
\caption{MAE$_x$ by number of basis functions across different datasets on basis learning task}
\label{fig:mae_x_basis}
\end{figure}
  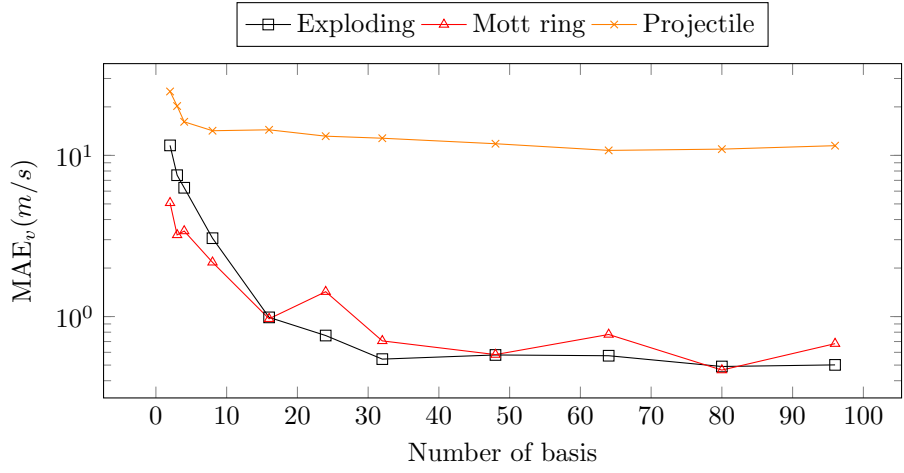
\begin{figure}[h]
\centering
\begin{tikzpicture}
\begin{axis}[
    legend style={at={(0.5,1.05)}, anchor=south},
    legend columns=-1, 
    width=\columnwidth,
    xlabel={Number of basis},
    ylabel={MAE$_v (m/s)$ },
    ymode=log,
    height=6cm
]
\addplot [
  color=black,
  mark=square
]coordinates {
(2, 1.1536e+01)
(3, 7.5359e+00)
(4, 6.2979e+00)
(8, 3.0641e+00)
(16, 9.8849e-01)
(24, 7.6281e-01)
(32, 5.4460e-01)
(48, 5.7786e-01)
(64, 5.7155e-01)
(80, 4.9002e-01)
(96, 5.0145e-01)
};
\addlegendentry{Exploding};


\addplot [
  color=red,
  mark=triangle
]coordinates {
(2, 5.0766e+00)
(3, 3.2103e+00)
(4, 3.3941e+00)
(8, 2.1710e+00)
(16, 9.6885e-01)
(24, 1.4264e+00)
(32, 7.0658e-01)
(48, 5.8100e-01)
(64, 7.7433e-01)
(80, 4.6563e-01)
(96, 6.7914e-01)
};
\addlegendentry{Mott ring};

\addplot [
  color=orange,
  mark=x
]coordinates {
(2, 2.4917e+01)
(3, 2.0239e+01)
(4, 1.6158e+01)
(8, 1.4224e+01)
(16, 1.4413e+01)
(24, 1.3159e+01)
(32, 1.2779e+01)
(48, 1.1812e+01)
(64, 1.0739e+01)
(80, 1.0936e+01)
(96, 1.1489e+01)

};
\addlegendentry{Projectile};
\end{axis}
\end{tikzpicture}
\caption{MAE$_v$ by number of basis functions across different datasets on basis learning task}
\label{fig:mae_v_basis}
\end{figure}

We found that Function Encoder can accurately reconstruct the space of state function with a small set of basis functions, as demonstrated by the R$^2$ score exceeding $99\%$ in most datasets in table \ref{tb:basis_r2}. We also found that the expressiveness of the Function Encoder is increasing with the number of basis functions but plateaus after $32$ bases, as shown in figures \ref{fig:mae_x_basis} and \ref{fig:mae_v_basis}.
}

\paragraph{Learning dynamic function}{
  Given a trained Function Encoder, where we could represent a function by a set of coefficients, we want to examine whether this set of coefficients is suitable for the task of dynamic learning.
  At each time step, we compute the state of the system by either equation \ref{eq:coef_inner} or \ref{eq:coef_lsq}.
  Then the state of the system is evolved to the subsequent time step by equation \ref{eq:discrete_dynamic} in the case of \emph{discrete time} dynamical system and equation \ref{eq:continuous_dynamic} in the case of \emph{continuous time} dynamical system.
  The state function at the future state is recovered using equation \ref{eq: reconstruct}.
  The reconstruction loss is then evaluated by comparing the evaluation of the future state function and the ground truth.

  \begin{table}[h]
  \begin{tabular}{
      >{\arraybackslash}p{1.25cm}
      >{\centering\arraybackslash}p{1cm}
      c
      c
    }
    \toprule
    &  & Train R$^2$ & Test R$^2$ \\
    Dataset & Basis ($K$) &  &  \\
    \midrule
    \multirow[t]{11}{*}{Exploding} & 2 & $\num{98.1098} \quad\hfill ( \pm \num{0.1578} )$ & $\num{98.5908} \quad\hfill ( \pm \num{0.0866} )$ \\
    & 4 & $\num{98.8048} \quad\hfill ( \pm \num{0.3080} )$ & $\num{99.3325} \quad\hfill ( \pm \num{0.0323} )$ \\
    & 8 & $\num{99.7885} \quad\hfill ( \pm \num{0.0552} )$ & $\num{99.8876} \quad\hfill ( \pm \num{0.0054} )$ \\
    & 16 & $\num{99.9890} \quad\hfill ( \pm \num{0.0083} )$ & $\num{99.9958} \quad\hfill ( \pm \num{0.0012} )$ \\
    & 32 & $\num{99.9982} \quad\hfill ( \pm \num{0.0002} )$ & $\num{99.9982} \quad\hfill ( \pm \num{0.0002} )$ \\
    & 64 & $\num{99.9980} \quad\hfill ( \pm \num{0.0006} )$ & $\num{99.9983} \quad\hfill ( \pm \num{0.0001} )$ \\
    & 96 & $\num{99.9984} \quad\hfill ( \pm \num{0.0005} )$ & $\num{99.9989} \quad\hfill ( \pm \num{0.0001} )$ \\
    \cline{1-4}
    \multirow[t]{11}{*}{Mott ring} & 2 & $\num{99.8808} \quad\hfill ( \pm \num{0.0116} )$ & $\num{99.9034} \quad\hfill ( \pm \num{0.0194} )$ \\
    & 4 & $\num{99.9436} \quad\hfill ( \pm \num{0.0364} )$ & $\num{99.9827} \quad\hfill ( \pm \num{0.0106} )$ \\
    & 8 & $\num{99.9699} \quad\hfill ( \pm \num{0.0170} )$ & $\num{99.9888} \quad\hfill ( \pm \num{0.0048} )$ \\
    & 16 & $\num{99.9849} \quad\hfill ( \pm \num{0.0121} )$ & $\num{99.9972} \quad\hfill ( \pm \num{0.0002} )$ \\
    & 32 & $\num{99.9941} \quad\hfill ( \pm \num{0.0049} )$ & $\num{99.9971} \quad\hfill ( \pm \num{0.0022} )$ \\
    & 64 & $\num{99.9968} \quad\hfill ( \pm \num{0.0014} )$ & $\num{99.9979} \quad\hfill ( \pm \num{0.0013} )$ \\
    & 96 & $\num{99.9973} \quad\hfill ( \pm \num{0.0021} )$ & $\num{99.9992} \quad\hfill ( \pm \num{0.0002} )$ \\
    \cline{1-4}
    \multirow[t]{11}{*}{Projectile} & 2 & $\num{89.6531} \quad\hfill ( \pm \num{5.2606} )$ & $\num{94.8480} \quad\hfill ( \pm \num{0.5477} )$ \\
    & 4 & $\num{97.2062} \quad\hfill ( \pm \num{0.7094} )$ & $\num{97.7648} \quad\hfill ( \pm \num{0.0890} )$ \\
    & 8 & $\num{97.7770} \quad\hfill ( \pm \num{0.2725} )$ & $\num{98.6154} \quad\hfill ( \pm \num{0.0234} )$ \\
    & 16 & $\num{98.5947} \quad\hfill ( \pm \num{0.1866} )$ & $\num{98.6641} \quad\hfill ( \pm \num{0.1498} )$ \\
    & 32 & $\num{98.9991} \quad\hfill ( \pm \num{0.1804} )$ & $\num{99.0932} \quad\hfill ( \pm \num{0.0822} )$ \\
    & 64 & $\num{98.8969} \quad\hfill ( \pm \num{0.4136} )$ & $\num{99.0991} \quad\hfill ( \pm \num{0.2358} )$ \\
    & 96 & $\num{98.9779} \quad\hfill ( \pm \num{0.2903} )$ & $\num{99.0927} \quad\hfill ( \pm \num{0.1057} )$ \\
    \cline{1-4}
    \bottomrule
  \end{tabular}
  \caption{Average and standard deviation of R$^2$ ($\mu \pm \sigma$) across different datasets and number of basis functions on dynamic learning task.}
  \label{tb:dynamic_r2}
\end{table}

  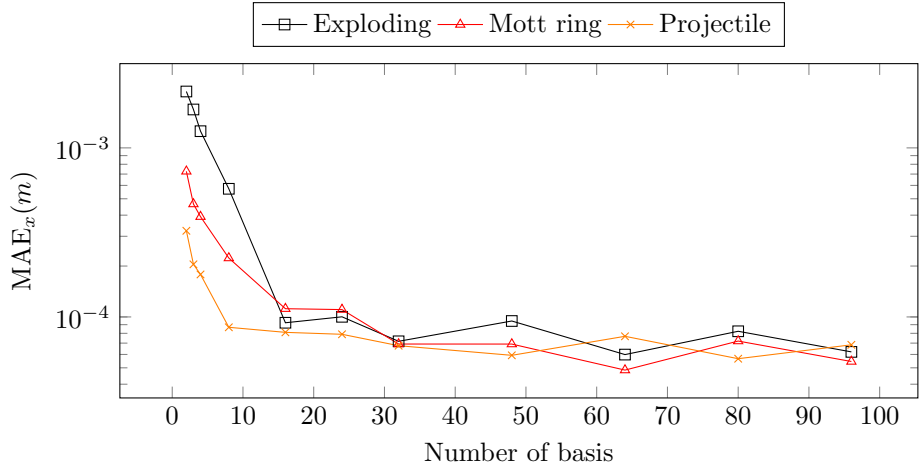
\begin{figure}[h]
\centering
\begin{tikzpicture}
\begin{axis}[
    legend style={at={(0.5,1.05)}, anchor=south},
    legend columns=-1, 
    width=\columnwidth,
    xlabel={Number of basis},
    ylabel={MAE$_x (m)$},
    ymode=log,
    height=6cm
]
\addplot [
  color=black,
  mark=square
]coordinates {
(2, 2.1558e-03)
(3, 1.6874e-03)
(4, 1.2556e-03)
(8, 5.7283e-04)
(16, 9.2371e-05)
(24, 1.0017e-04)
(32, 7.1766e-05)
(48, 9.4612e-05)
(64, 5.9886e-05)
(80, 8.2261e-05)
(96, 6.2122e-05)
};
\addlegendentry{Exploding};


\addplot [
  color=red,
  mark=triangle
]coordinates {
(2, 7.2675e-04)
(3, 4.6549e-04)
(4, 3.9193e-04)
(8, 2.2304e-04)
(16, 1.1186e-04)
(24, 1.1053e-04)
(32, 6.9149e-05)
(48, 6.9112e-05)
(64, 4.8486e-05)
(80, 7.2001e-05)
(96, 5.4501e-05)
};
\addlegendentry{Mott ring};

\addplot [
  color=orange,
  mark=x
]coordinates {
(2, 3.2334e-04)
(3, 2.0477e-04)
(4, 1.7840e-04)
(8, 8.6839e-05)
(16, 8.1058e-05)
(24, 7.8958e-05)
(32, 6.7600e-05)
(48, 5.9290e-05)
(64, 7.6808e-05)
(80, 5.6620e-05)
(96, 6.8410e-05)
};
\addlegendentry{Projectile};
\end{axis}
\end{tikzpicture}
\caption{MAE$_x$ by number of basis functions across different datasets on dynamic learning task}
\label{fig:mae_x_dynamic}
\end{figure}
  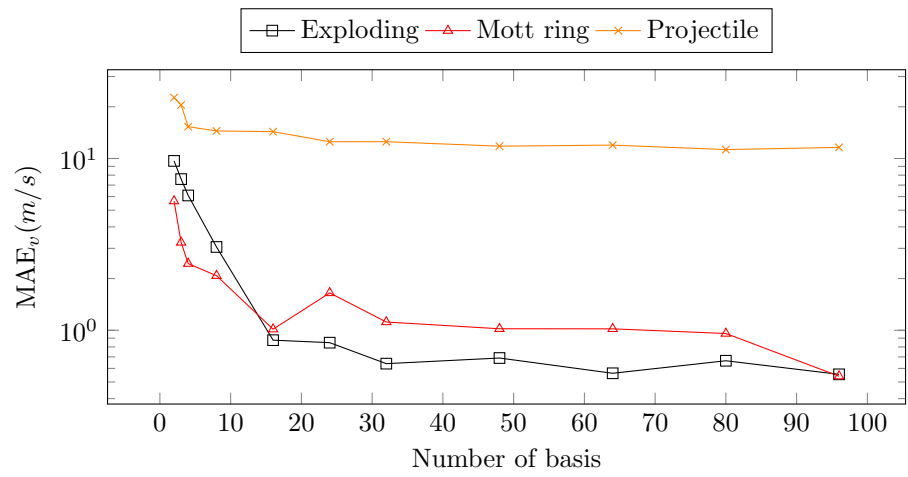
\begin{figure}[h]
\centering
\begin{tikzpicture}
\begin{axis}[
    legend style={at={(0.5,1.05)}, anchor=south},
    legend columns=-1, 
    width=\columnwidth,
    xlabel={Number of basis},
    ylabel={MAE$_v (m/s)$ },
    ymode=log,
    height=6cm
]
\addplot [
  color=black,
  mark=square
]coordinates {
(2, 9.6819e+00)
(3, 7.5878e+00)
(4, 6.0923e+00)
(8, 3.0550e+00)
(16, 8.7598e-01)
(24, 8.4701e-01)
(32, 6.3933e-01)
(48, 6.8955e-01)
(64, 5.6196e-01)
(80, 6.6471e-01)
(96, 5.5409e-01)
};
\addlegendentry{Exploding};


\addplot [
  color=red,
  mark=triangle
]coordinates {
(2, 5.6501e+00)
(3, 3.2461e+00)
(4, 2.4433e+00)
(8, 2.0784e+00)
(16, 1.0147e+00)
(24, 1.6503e+00)
(32, 1.1173e+00)
(48, 1.0207e+00)
(64, 1.0196e+00)
(80, 9.5774e-01)
(96, 5.4047e-01)
};
\addlegendentry{Mott ring};

\addplot [
  color=orange,
  mark=x
]coordinates {
(2, 2.2622e+01)
(3, 2.0498e+01)
(4, 1.5316e+01)
(8, 1.4472e+01)
(16, 1.4326e+01)
(24, 1.2519e+01)
(32, 1.2524e+01)
(48, 1.1797e+01)
(64, 1.1964e+01)
(80, 1.1270e+01)
(96, 1.1604e+01)
};
\addlegendentry{Projectile};

\end{axis}
\end{tikzpicture}
\caption{MAE$_v$ by number of basis functions across different datasets on dynamic learning task}
\label{fig:mae_v_dynamic}
\end{figure}

  We found that the future state is accurately reconstructed as the R$^2$ score is exceeded $99\%$ in most datasets, as shown in table \ref{tb:dynamic_r2}. Similar to the results from the basis learning task, we also find that the accuracy of the model is increasing as we increase the number of basis functions, but in most cases plateaus after $32$ bases as shown in \ref{fig:mae_x_dynamic} and \ref{fig:mae_v_dynamic}.
}


\section{Conclusion and discussion}
\label{sec:conclude}


In this work, borrowing language from Continuum Mechanics, we proposed a scalable, non-intrusive reduced-order modeling framework for Lagrangian simulations. We formulated the system state as a function of the reference configuration. This approach departs from traditional graph-based methods that treat the system as discrete particles. As a result, we are able to reduce the degrees of freedom by several orders of magnitude. 
In comparison to other methods that also represent the state as a function, our approach parameterizes the space of functions as a linear space spanned by a set of basis functions. In contrast, previous methods represent this space as a nonlinear manifold, which involves more complex parameterizations. By choosing a linear space, we can access the state of the system through straightforward projection, avoiding additional complexities associated with nonlinear manifolds.

From a representation learning perspective, the Function Encoder could be understood as an autoencoder for irregular point cloud data, where the encoder maps a discretized function to a latent vector, and the decoder is a linear combination of learned basis functions. This form formulation admits a clear interpretation that the latent vector corresponds to the set of coefficients of basis functions that span the function space.

More broadly, our method is a conceptual bridge between Proper Orthogonal Decomposition, which is a classical method of reduced-order modeling, and modern deep learning methods. Proper Orthogonal Decomposition scales poorly with the spatial resolution and is often limited to the Eulerian setting due to its reliance on Principal Components Analysis to decompose the snapshot matrix into spatial modes. In contrast, our method replaces the discretized spatial modes with a learned basis function and could be evaluated at an arbitrary location. This parameterization allows our approach to overcome the scalability problem and extends it to the Lagrangian system. 


Furthermore, by having access to the material and spatial gradient and efficient evaluation of the system's state, our method offers avenues to incorporate physics before developing physics-informed ROM. In future works, we would like to examine the performance of models when imposing additional physics constraints such as conservation of mass, momentum, and energy. The separation between representation and dynamics also offers an avenue to incorporate the principle of least action by choosing a specialized architecture for the dynamic model, such as Hamiltonian or Lagrangian Neural Networks.

However, one of the main obstacles we would like to address in future work is the limitation to a single reference configuration.
While our method applies to the case of Eulerian simulation, where the computation domain is fixed, or Lagrangian simulation with a single reference configuration, this limitation precludes application to realistic use cases where multiple simulations have different reference configurations and highly irregular geometry.


\printbibliography

\clearpage
\appendix
\section{Table of notations}

\begin{table}[H]
  \begin{tabularx}{\columnwidth}{p{2cm}p{6cm}}
    \toprule
    \textbf{Symbol} & \textbf{Notation} \\ \midrule
    $\Omega, \Omega_0$ &
    Reference and current configuration\\

    $\inner{\cdot}{\cdot}, \inner{\cdot}{\cdot}_\Omega$ &
    Inner product, and inner product defined on some space $\Omega$\\

    $\mathcal{F},\mathcal{S}$
    & Function space\\

    $f, x, u, v, \cdots$
    & Member of a function space\\

    $\{(\mathbf{x}_i, \mathbf{u}_i)\}_{\mathbf{x}_i \in \Omega}$
    & Discretized points of a vector field $u(x)$ defined over the domain $\Omega$
    \\ \bottomrule
  \end{tabularx}
\end{table}

\section{Table of acronyms}

\begin{table}[H]
  \begin{tabularx}{\columnwidth}{p{2cm}p{6cm}}
    \toprule
    \textbf{Abbreviation} & \textbf{Definition} \\ \midrule

    ROM &
    Reduced-Order Modeling\\

    MMOD &
    Micrometeoroid and Orbital Debris\\

    SPH & 
    Smoothed Particle Hydrodynamic\\

    MPM &
    Material Point Method\\

    PDE &
    Partial Differential Equations\\

    \\ \bottomrule
  \end{tabularx}
\end{table}

\section{Continuum Mechanics}
\label{apdx:continuum_mech}
\begin{definition}[\emph{Configuration}]
  Consider a body $\mathcal{B}$ in three-dimensional space $\mathbb{R}^3$; $\mathcal{B}$ can be represented by a collection of particles called material points. Given a coordinate system, at each instant in time, each point can be represented by a unique vector $\mathbf{x}$, and the object can be represented by a set of vectors $\Omega = \{ \mathbf{x}^{(i)}\}_{i=1\cdots N}$. The set of vectors $\Omega$ is called \emph{configuration} of the object.

\end{definition}

\begin{definition}[\emph{Reference configuration} and \emph{Lagrangian coordinates}]
  The configuration that is corresponding to the material points at time $t=0$ is called \emph{reference configuration} and is often denoted as $\Omega_0 = \{\mathbf{X}^{(i)}\}_{i=1,\cdots N}$. Where the coordinates of the $i^{th}$ material point in the reference configuration, denoted by $\mathbf{X}^{(i)}$, are called \emph{material coordinates} or \emph{Lagrangian coordinates}.
\end{definition}

\begin{definition}[\emph{Current configuration} and \emph{Eulerian coordinates}]
  After time $t$ under deformation, the object occupies another configuration, which is called \emph{current configuration}. The configuration is represented by another set of vectors $\Omega = \{\mathbf{x}^{(i)}\}_{i=1\cdots N}$, where $\mathbf{x}^{(i)}$ is the coordinates of the $i^{th}$ material points. These coordinates are called \emph{spatial coordinates} or \emph{Eulerian coordinates}.
\end{definition}

\begin{definition}[\emph{Deformation mapping} $f: \Omega_0 \rightarrow \Omega$]
  Deformation mapping of the body $\mathcal{B}$ from $\Omega_0$ to $\Omega$. The deformation mapping $f(\mathbf{X}, t)$ takes the position vector $\mathbf{X}$ from the reference configuration and places the same material point into the current configuration $\mathbf{x}=f(\mathbf{X}, t)$. The inverse mapping $f^{-1}: \Omega \rightarrow\Omega_0$ take the position vector $\mathbf{x}$ from current configuration back to reference configuration $\mathbf{X}=f^{-1}(\mathbf{x}, t)$.
\end{definition}

\section{Hilbert space and axioms of an inner product function}
\label{apdx:inner_product}
\begin{definition}[\emph{Hilbert Space} and \emph{Inner product}]
  is a complete vector space $\mathcal{X}$ equipped with an \emph{inner product}. An inner product, which is a function that takes two members of the function space and maps to the real number line, satisfies four axioms in appendix \ref{apdx:inner_product}.
\end{definition}

  \underline{\bf Inner product} $\inner{\cdot, \cdot} : \mathcal{X}\times\mathcal{X}\rightarrow \mathbb{R}$, such that $\forall x, y, z \in \mathcal{X}; \forall \alpha \in \mathbb{R}$

  \begin{equation*}
    \begin{aligned}
      &\text{\bf(IP1)} &\quad \inner{x+y}{z}       &= \inner{x}{z} + \inner{y}{z} \\
      &\text{\bf(IP2)} &\quad \inner{\alpha x}{y}  &= \alpha\inner{x}{y}\\
      &\text{\bf(IP3)} &\quad \inner{x}{y}         &= \inner{y}{x} \footnote{(For real vector space)}\\
      &\text{\bf(IP4)} &\quad \inner{x}{x}         &\geq 0\\
      &&\inner{x}{x} &= 0 \iff x = 0
    \end{aligned}
  \end{equation*}

%

\section{Algorithms}
\label{apdx:fe_algo}
\begin{algorithm}[H]
  \caption{Function Encoder’s algorithm}\label{alg:fe}

  \begin{algorithmic}[1]
    \Procedure{FunctionEncoder}{$\text{max\_step}, \mathcal{D}$}
    \LComment{$- \mathcal{D}=\{S_j\}_{j=1}^T$: collection of snapshot of states of the dynamical system. \\- max\_step: number of gradient steps.} 
    \State $\Theta \gets$ \texttt{Initialized()}
    \Comment{Parameters of the set of basis functions.}
    \State $ i \gets 1$
    \While{$ i \le \text{max\_step}$}
      \LComment{$S_j$: set of discretization point of $s(X, t):\forall X \in \Omega_0$}
      \State $S_j \defeq \{(X^{(i)}, s^{(i)}_j)\}_{i=1}^N \gets \texttt{Sample}(\mathcal{D})$

      \LComment{$S^e_j, S^q_j$ are disjoint sets of example and query points.}

      \State $S^e_j \defeq (X^e, s^e_j) \gets \texttt{Sample}(S_j)$
      \LComment{$X^e = \{ X^{(i)}\}_{i=1}^M \\ s^{e}_j = \{ s(X, t_j): X \in X^e \}$}
      \State $S^q_j \defeq (X^q, s^q_j) \gets \texttt{Sample}(S_j)$
      \LComment{$X^q = \{ X^{(i)}\}_{i=1}^M \\ s^{q}_j = \{ s(X, t_j): X \in X^q \}$}

      \State $\mathbf{\hat{c}}_j\gets \texttt{Encode}(X^e, s^e_j)$
      \Comment{Coefficient in Hilbert space}
      \State $\hat{s}^q_j \gets \texttt{Decode}(X^q, \mathbf{\hat{c}}_j)$
      \Comment{Reconstruction of the vector field}
      \State $\mathcal{L}(\Theta) = \texttt{MSE}(s^q_j, \hat{s}^q_j)$
      \Comment{Loss function}
      \State $\Theta \gets \Theta + \alpha \nabla\mathcal{L}$
      \Comment{Learning rate $\alpha$}

      \State $i \gets i + 1$
    \EndWhile
    \State \textbf{return} $\Theta$
    \EndProcedure
  \end{algorithmic}
  \begin{algorithmic}[2]
    \Procedure{Encode}{$X, s, \Theta$}
      \State $$
        \Phi \gets \begin{bmatrix}
          \phi_1(X_1; \theta_1) & \phi_2(X_1; \theta_2) & \cdots & \phi_K(X_1;\theta_K)\\
          \phi_1(X_2; \theta_1) & \phi_2(X_2; \theta_2) & \cdots & \phi_K(X_2;\theta_K)\\
          \vdots & \vdots & \ddots & \vdots\\
          \phi_1(X_M; \theta_1) & \phi_2(X_M; \theta_2) & \cdots & \phi_K(X_M; \theta_K)\\
        \end{bmatrix}
      $$

      \State \textbf{return} $(\Phi^\top\Phi)^{-1}\Phi^\top s$
    \EndProcedure
  \end{algorithmic}
  \begin{algorithmic}[3]
    \Procedure{Decode}{$X, \mathbf{c}, \Theta$}
      \State \textbf{return} $\sum_{k=1}^K{c_k\phi_k(X;\theta_k)}$
    \EndProcedure
  \end{algorithmic}
\end{algorithm}

\section{Further results}
\label{apdx:more_result}

\subsection{On baseline dataset}
\begin{table*}
\resizebox{\textwidth}{!}{
\begin{tabular}{lcrllll}
\hline
\multirow{2}{*}{Dataset}                                                  & \multirow{2}{*}{\begin{tabular}[c]{@{}c@{}}Sample\\ size\end{tabular}} & \multicolumn{1}{l}{\multirow{2}{*}{Basis}} & \multicolumn{2}{c}{Train}                                                                                                    & \multicolumn{2}{c}{Test}                                                                                \\ \cline{4-7} 
                                                                          &                                                                        & \multicolumn{1}{l}{}                       & \multicolumn{1}{c}{In-t}                           & \multicolumn{1}{c|}{Out-t}                                              & \multicolumn{1}{c}{In-t}                           & \multicolumn{1}{c}{Out-t}                          \\ \hline
\multirow{12}{*}{\begin{tabular}[c]{@{}l@{}}Navier\\ Stokes\end{tabular}} & \multirow{4}{*}{5\%}                                                   & 32                                         & $\num{4.667e-03} \quad\hfill ( \pm\num{4.8e-05} )$ & \multicolumn{1}{l|}{$\num{7.359e-03} \quad\hfill ( \pm\num{1.1e-04} )$} & $\num{5.403e-03} \quad\hfill ( \pm\num{1.0e-04} )$ & $\num{7.851e-03} \quad\hfill ( \pm\num{2.1e-04} )$ \\
                                                                          &                                                                        & 64                                         & $\num{2.351e-03} \quad\hfill ( \pm\num{2.9e-05} )$ & \multicolumn{1}{l|}{$\num{5.594e-03} \quad\hfill ( \pm\num{5.1e-05} )$} & $\num{3.213e-03} \quad\hfill ( \pm\num{3.7e-05} )$ & $\num{6.925e-03} \quad\hfill ( \pm\num{5.8e-05} )$ \\
                                                                          &                                                                        & 96                                         & $\num{3.105e-03} \quad\hfill ( \pm\num{3.4e-05} )$ & \multicolumn{1}{l|}{$\num{7.796e-03} \quad\hfill ( \pm\num{1.0e-04} )$} & $\num{3.896e-03} \quad\hfill ( \pm\num{1.9e-05} )$ & $\num{8.834e-03} \quad\hfill ( \pm\num{5.7e-05} )$ \\
                                                                          &                                                                        & 128                                        & $\num{3.578e-03} \quad\hfill ( \pm\num{2.0e-05} )$ & \multicolumn{1}{l|}{$\num{8.494e-03} \quad\hfill ( \pm\num{6.1e-05} )$} & $\num{4.592e-03} \quad\hfill ( \pm\num{1.8e-05} )$ & $\num{1.002e-02} \quad\hfill ( \pm\num{1.6e-04} )$ \\ \cline{2-7} 
                                                                          & \multirow{4}{*}{25\%}                                                  & 32                                         & $\num{4.379e-03} \quad\hfill ( \pm\num{1.5e-05} )$ & \multicolumn{1}{l|}{$\num{7.120e-03} \quad\hfill ( \pm\num{4.2e-05} )$} & $\num{5.026e-03} \quad\hfill ( \pm\num{1.4e-05} )$ & $\num{7.365e-03} \quad\hfill ( \pm\num{7.3e-05} )$ \\
                                                                          &                                                                        & 64                                         & $\num{2.163e-03} \quad\hfill ( \pm\num{1.2e-05} )$ & \multicolumn{1}{l|}{$\num{5.431e-03} \quad\hfill ( \pm\num{2.1e-05} )$} & $\num{2.988e-03} \quad\hfill ( \pm\num{1.5e-05} )$ & $\num{6.644e-03} \quad\hfill ( \pm\num{4.3e-05} )$ \\
                                                                          &                                                                        & 96                                         & $\num{2.870e-03} \quad\hfill ( \pm\num{6.1e-06} )$ & \multicolumn{1}{l|}{$\num{7.593e-03} \quad\hfill ( \pm\num{4.5e-05} )$} & $\num{3.624e-03} \quad\hfill ( \pm\num{1.4e-05} )$ & $\num{8.599e-03} \quad\hfill ( \pm\num{2.2e-05} )$ \\
                                                                          &                                                                        & 128                                        & $\num{3.258e-03} \quad\hfill ( \pm\num{7.9e-06} )$ & \multicolumn{1}{l|}{$\num{8.142e-03} \quad\hfill ( \pm\num{3.9e-05} )$} & $\num{4.296e-03} \quad\hfill ( \pm\num{5.6e-06} )$ & $\num{9.759e-03} \quad\hfill ( \pm\num{8.2e-05} )$ \\ \cline{2-7} 
                                                                          & \multirow{4}{*}{50\%}                                                  & 32                                         & $\num{4.341e-03} \quad\hfill ( \pm\num{8.3e-06} )$ & \multicolumn{1}{l|}{$\num{7.105e-03} \quad\hfill ( \pm\num{4.5e-05} )$} & $\num{5.003e-03} \quad\hfill ( \pm\num{7.7e-06} )$ & $\num{7.408e-03} \quad\hfill ( \pm\num{4.2e-05} )$ \\
                                                                          &                                                                        & 64                                         & $\num{2.157e-03} \quad\hfill ( \pm\num{1.2e-05} )$ & \multicolumn{1}{l|}{$\num{5.446e-03} \quad\hfill ( \pm\num{4.5e-05} )$} & $\num{2.954e-03} \quad\hfill ( \pm\num{4.5e-06} )$ & $\num{6.656e-03} \quad\hfill ( \pm\num{3.4e-05} )$ \\
                                                                          &                                                                        & 96                                         & $\num{2.854e-03} \quad\hfill ( \pm\num{6.7e-06} )$ & \multicolumn{1}{l|}{$\num{7.528e-03} \quad\hfill ( \pm\num{3.2e-05} )$} & $\num{3.618e-03} \quad\hfill ( \pm\num{1.1e-05} )$ & $\num{8.568e-03} \quad\hfill ( \pm\num{2.8e-05} )$ \\
                                                                          &                                                                        & 128                                        & $\num{3.256e-03} \quad\hfill ( \pm\num{1.5e-05} )$ & \multicolumn{1}{l|}{$\num{8.193e-03} \quad\hfill ( \pm\num{7.0e-05} )$} & $\num{4.280e-03} \quad\hfill ( \pm\num{8.3e-06} )$ & $\num{9.739e-03} \quad\hfill ( \pm\num{5.6e-05} )$ \\ \hline
\multirow{12}{*}{Wave}                                                    & \multirow{4}{*}{5\%}                                                   & 32                                         & $\num{8.567e-03} \quad\hfill ( \pm\num{3.6e-05} )$ & \multicolumn{1}{l|}{$\num{7.952e-03} \quad\hfill ( \pm\num{5.7e-05} )$} & $\num{1.404e-02} \quad\hfill ( \pm\num{5.0e-05} )$ & $\num{1.490e-02} \quad\hfill ( \pm\num{3.2e-04} )$ \\
                                                                          &                                                                        & 64                                         & $\num{3.396e-04} \quad\hfill ( \pm\num{1.2e-05} )$ & \multicolumn{1}{l|}{$\num{1.350e-03} \quad\hfill ( \pm\num{3.1e-05} )$} & $\num{1.590e-03} \quad\hfill ( \pm\num{2.3e-05} )$ & $\num{7.449e-03} \quad\hfill ( \pm\num{1.1e-04} )$ \\
                                                                          &                                                                        & 96                                         & $\num{7.922e-05} \quad\hfill ( \pm\num{4.5e-06} )$ & \multicolumn{1}{l|}{$\num{1.036e-03} \quad\hfill ( \pm\num{1.3e-05} )$} & $\num{9.041e-04} \quad\hfill ( \pm\num{5.2e-06} )$ & $\num{6.645e-03} \quad\hfill ( \pm\num{2.3e-05} )$ \\
                                                                          &                                                                        & 128                                        & $\num{9.440e-05} \quad\hfill ( \pm\num{2.6e-06} )$ & \multicolumn{1}{l|}{$\num{1.016e-03} \quad\hfill ( \pm\num{1.4e-05} )$} & $\num{8.906e-04} \quad\hfill ( \pm\num{2.6e-05} )$ & $\num{5.772e-03} \quad\hfill ( \pm\num{1.4e-04} )$ \\ \cline{2-7} 
                                                                          & \multirow{4}{*}{25\%}                                                  & 32                                         & $\num{8.131e-03} \quad\hfill ( \pm\num{2.6e-05} )$ & \multicolumn{1}{l|}{$\num{7.541e-03} \quad\hfill ( \pm\num{3.0e-05} )$} & $\num{1.332e-02} \quad\hfill ( \pm\num{2.6e-05} )$ & $\num{1.450e-02} \quad\hfill ( \pm\num{1.1e-04} )$ \\
                                                                          &                                                                        & 64                                         & $\num{2.653e-04} \quad\hfill ( \pm\num{2.3e-06} )$ & \multicolumn{1}{l|}{$\num{1.243e-03} \quad\hfill ( \pm\num{8.9e-06} )$} & $\num{1.435e-03} \quad\hfill ( \pm\num{3.0e-05} )$ & $\num{7.127e-03} \quad\hfill ( \pm\num{1.5e-04} )$ \\
                                                                          &                                                                        & 96                                         & $\num{8.141e-05} \quad\hfill ( \pm\num{7.6e-06} )$ & \multicolumn{1}{l|}{$\num{1.054e-03} \quad\hfill ( \pm\num{3.1e-05} )$} & $\num{9.017e-04} \quad\hfill ( \pm\num{6.3e-06} )$ & $\num{6.672e-03} \quad\hfill ( \pm\num{5.0e-05} )$ \\
                                                                          &                                                                        & 128                                        & $\num{8.877e-05} \quad\hfill ( \pm\num{1.9e-06} )$ & \multicolumn{1}{l|}{$\num{1.001e-03} \quad\hfill ( \pm\num{1.8e-05} )$} & $\num{8.650e-04} \quad\hfill ( \pm\num{1.9e-05} )$ & $\num{5.657e-03} \quad\hfill ( \pm\num{9.3e-05} )$ \\ \cline{2-7} 
                                                                          & \multirow{4}{*}{50\%}                                                  & 32                                         & $\num{8.088e-03} \quad\hfill ( \pm\num{4.3e-06} )$ & \multicolumn{1}{l|}{$\num{7.498e-03} \quad\hfill ( \pm\num{2.0e-05} )$} & $\num{1.325e-02} \quad\hfill ( \pm\num{1.6e-05} )$ & $\num{1.430e-02} \quad\hfill ( \pm\num{1.2e-04} )$ \\
                                                                          &                                                                        & 64                                         & $\num{2.616e-04} \quad\hfill ( \pm\num{3.7e-06} )$ & \multicolumn{1}{l|}{$\num{1.260e-03} \quad\hfill ( \pm\num{1.9e-05} )$} & $\num{1.420e-03} \quad\hfill ( \pm\num{8.2e-06} )$ & $\num{7.100e-03} \quad\hfill ( \pm\num{4.9e-05} )$ \\
                                                                          &                                                                        & 96                                         & $\num{7.952e-05} \quad\hfill ( \pm\num{1.1e-05} )$ & \multicolumn{1}{l|}{$\num{1.053e-03} \quad\hfill ( \pm\num{4.6e-05} )$} & $\num{8.819e-04} \quad\hfill ( \pm\num{6.7e-06} )$ & $\num{6.530e-03} \quad\hfill ( \pm\num{4.8e-05} )$ \\
                                                                          &                                                                        & 128                                        & $\num{8.940e-05} \quad\hfill ( \pm\num{2.6e-06} )$ & \multicolumn{1}{l|}{$\num{1.001e-03} \quad\hfill ( \pm\num{6.5e-06} )$} & $\num{8.724e-04} \quad\hfill ( \pm\num{1.2e-05} )$ & $\num{5.693e-03} \quad\hfill ( \pm\num{6.5e-05} )$ \\ \hline
\end{tabular}
}
\caption{OnlyDense performance on baseline dataset across different setting.}
\label{tb:baseline_full}
\end{table*}

\subsection{Basis}
\begin{table*}
  \begin{tabular}{
      |l
      |c
      |c
      |c|
    }
    \hline
    &  & Train R$^2$ & Test R$^2$ \\
    Dataset & Basis ($K$) &  &  \\
    \hline
    \multirow[t]{11}{*}{Exploding} & 2 & $\num{98.3416} \quad\hfill ( \pm \num{0.2419} )$ & $\num{98.6247} \quad\hfill ( \pm \num{0.0188} )$ \\
    & 3 & $\num{98.8587} \quad\hfill ( \pm \num{0.1114} )$ & $\num{99.0752} \quad\hfill ( \pm \num{0.0959} )$ \\
    & 4 & $\num{99.2290} \quad\hfill ( \pm \num{0.1218} )$ & $\num{99.2763} \quad\hfill ( \pm \num{0.0358} )$ \\
    & 8 & $\num{99.8650} \quad\hfill ( \pm \num{0.0255} )$ & $\num{99.8882} \quad\hfill ( \pm \num{0.0043} )$ \\
    & 16 & $\num{99.9831} \quad\hfill ( \pm \num{0.0098} )$ & $\num{99.9950} \quad\hfill ( \pm \num{0.0020} )$ \\
    & 24 & $\num{99.9960} \quad\hfill ( \pm \num{0.0006} )$ & $\num{99.9970} \quad\hfill ( \pm \num{0.0002} )$ \\
    & 32 & $\num{99.9983} \quad\hfill ( \pm \num{0.0006} )$ & $\num{99.9991} \quad\hfill ( \pm \num{0.0001} )$ \\
    & 48 & $\num{99.9990} \quad\hfill ( \pm \num{0.0004} )$ & $\num{99.9994} \quad\hfill ( \pm \num{0.0001} )$ \\
    & 64 & $\num{99.9991} \quad\hfill ( \pm \num{0.0003} )$ & $\num{99.9992} \quad\hfill ( \pm \num{0.0001} )$ \\
    & 80 & $\num{99.9993} \quad\hfill ( \pm \num{0.0001} )$ & $\num{99.9995} \quad\hfill ( \pm \num{0.0002} )$ \\
    & 96 & $\num{99.9993} \quad\hfill ( \pm \num{0.0002} )$ & $\num{99.9995} \quad\hfill ( \pm \num{0.0001} )$ \\
    \cline{1-4}
    \multirow[t]{11}{*}{Mott ring} & 2 & $\num{99.7917} \quad\hfill ( \pm \num{0.0994} )$ & $\num{99.9107} \quad\hfill ( \pm \num{0.0104} )$ \\
    & 3 & $\num{99.9301} \quad\hfill ( \pm \num{0.0340} )$ & $\num{99.9649} \quad\hfill ( \pm \num{0.0180} )$ \\
    & 4 & $\num{99.9446} \quad\hfill ( \pm \num{0.0281} )$ & $\num{99.9678} \quad\hfill ( \pm \num{0.0123} )$ \\
    & 8 & $\num{99.9667} \quad\hfill ( \pm \num{0.0124} )$ & $\num{99.9899} \quad\hfill ( \pm \num{0.0025} )$ \\
    & 16 & $\num{99.9886} \quad\hfill ( \pm \num{0.0120} )$ & $\num{99.9973} \quad\hfill ( \pm \num{0.0002} )$ \\
    & 24 & $\num{99.9933} \quad\hfill ( \pm \num{0.0053} )$ & $\num{99.9952} \quad\hfill ( \pm \num{0.0024} )$ \\
    & 32 & $\num{99.9878} \quad\hfill ( \pm \num{0.0066} )$ & $\num{99.9985} \quad\hfill ( \pm \num{0.0002} )$ \\
    & 48 & $\num{99.9942} \quad\hfill ( \pm \num{0.0032} )$ & $\num{99.9990} \quad\hfill ( \pm \num{0.0001} )$ \\
    & 64 & $\num{99.9968} \quad\hfill ( \pm \num{0.0007} )$ & $\num{99.9986} \quad\hfill ( \pm \num{0.0011} )$ \\
    & 80 & $\num{99.9970} \quad\hfill ( \pm \num{0.0025} )$ & $\num{99.9995} \quad\hfill ( \pm \num{0.0000} )$ \\
    & 96 & $\num{99.9962} \quad\hfill ( \pm \num{0.0034} )$ & $\num{99.9990} \quad\hfill ( \pm \num{0.0006} )$ \\
    \cline{1-4}
    \multirow[t]{11}{*}{Projectile} & 2 & $\num{88.7186} \quad\hfill ( \pm \num{5.1813} )$ & $\num{94.8356} \quad\hfill ( \pm \num{1.7446} )$ \\
    & 3 & $\num{95.8493} \quad\hfill ( \pm \num{0.6640} )$ & $\num{96.5606} \quad\hfill ( \pm \num{0.1391} )$ \\
    & 4 & $\num{97.6520} \quad\hfill ( \pm \num{0.3678} )$ & $\num{97.9315} \quad\hfill ( \pm \num{0.1017} )$ \\
    & 8 & $\num{98.5252} \quad\hfill ( \pm \num{0.0934} )$ & $\num{98.6370} \quad\hfill ( \pm \num{0.0427} )$ \\
    & 16 & $\num{98.6602} \quad\hfill ( \pm \num{0.2050} )$ & $\num{98.6682} \quad\hfill ( \pm \num{0.0859} )$ \\
    & 24 & $\num{98.5937} \quad\hfill ( \pm \num{0.2881} )$ & $\num{99.0256} \quad\hfill ( \pm \num{0.0212} )$ \\
    & 32 & $\num{99.0982} \quad\hfill ( \pm \num{0.0516} )$ & $\num{99.0388} \quad\hfill ( \pm \num{0.0930} )$ \\
    & 48 & $\num{99.1662} \quad\hfill ( \pm \num{0.1677} )$ & $\num{99.2152} \quad\hfill ( \pm \num{0.0646} )$ \\
    & 64 & $\num{98.8312} \quad\hfill ( \pm \num{0.4004} )$ & $\num{99.1914} \quad\hfill ( \pm \num{0.1772} )$ \\
    & 80 & $\num{98.8092} \quad\hfill ( \pm \num{0.1794} )$ & $\num{99.2711} \quad\hfill ( \pm \num{0.0534} )$ \\
    & 96 & $\num{98.8706} \quad\hfill ( \pm \num{0.2491} )$ & $\num{99.1768} \quad\hfill ( \pm \num{0.1014} )$ \\
    \cline{1-4}
    \bottomrule
  \end{tabular}
  \caption{Average and standard deviation of R$^2$ ($\mu \pm \sigma$) across different datasets and number of basis functions on basis learning task.}
  \label{tb:full_r2_basis}
\end{table*}

\begin{table*}
  \begin{tabular}{
      |l
      |c
      |c
      |c
      |c
      |c|
    }
    \hline
    &  & Train MSE$_x$ & Test MSE$_x$ & Train MSE$_v$ & Test MSE$_v$ \\
    Dataset & Basis ($K$) &  &  &  &  \\
    \hline
    \multirow[t]{11}{*}{Exploding} & 2 & $\num{9.6e-06} \quad\hfill ( \pm\num{5e-06} )$ & $\num{2.7e-05} \quad\hfill ( \pm\num{5e-06} )$ & $\num{3.0e+02} \quad\hfill ( \pm\num{6e+01} )$ & $\num{3.6e+02} \quad\hfill ( \pm\num{4e+01} )$ \\
    & 3 & $\num{1.0e-05} \quad\hfill ( \pm\num{5e-06} )$ & $\num{1.2e-05} \quad\hfill ( \pm\num{6e-06} )$ & $\num{1.8e+02} \quad\hfill ( \pm\num{1e+01} )$ & $\num{1.8e+02} \quad\hfill ( \pm\num{1e+01} )$ \\
    & 4 & $\num{2.7e-06} \quad\hfill ( \pm\num{3e-06} )$ & $\num{4.0e-06} \quad\hfill ( \pm\num{5e-06} )$ & $\num{1.5e+02} \quad\hfill ( \pm\num{1e+01} )$ & $\num{1.5e+02} \quad\hfill ( \pm\num{8e+00} )$ \\
    & 8 & $\num{1.5e-06} \quad\hfill ( \pm\num{6e-07} )$ & $\num{1.5e-06} \quad\hfill ( \pm\num{4e-07} )$ & $\num{3.5e+01} \quad\hfill ( \pm\num{3e+00} )$ & $\num{3.0e+01} \quad\hfill ( \pm\num{2e+00} )$ \\
    & 16 & $\num{9.0e-08} \quad\hfill ( \pm\num{2e-08} )$ & $\num{5.0e-08} \quad\hfill ( \pm\num{3e-08} )$ & $\num{4.4e+00} \quad\hfill ( \pm\num{7e-01} )$ & $\num{2.6e+00} \quad\hfill ( \pm\num{1e+00} )$ \\
    & 24 & $\num{2.6e-08} \quad\hfill ( \pm\num{6e-09} )$ & $\num{3.2e-08} \quad\hfill ( \pm\num{1e-08} )$ & $\num{2.4e+00} \quad\hfill ( \pm\num{5e-01} )$ & $\num{2.1e+00} \quad\hfill ( \pm\num{5e-01} )$ \\
    & 32 & $\num{3.5e-09} \quad\hfill ( \pm\num{2e-09} )$ & $\num{2.8e-09} \quad\hfill ( \pm\num{6e-10} )$ & $\num{8.7e-01} \quad\hfill ( \pm\num{4e-01} )$ & $\num{8.0e-01} \quad\hfill ( \pm\num{4e-01} )$ \\
    & 48 & $\num{3.7e-09} \quad\hfill ( \pm\num{1e-09} )$ & $\num{3.7e-09} \quad\hfill ( \pm\num{1e-09} )$ & $\num{6.3e-01} \quad\hfill ( \pm\num{2e-01} )$ & $\num{8.8e-01} \quad\hfill ( \pm\num{3e-01} )$ \\
    & 64 & $\num{3.8e-09} \quad\hfill ( \pm\num{5e-10} )$ & $\num{4.0e-09} \quad\hfill ( \pm\num{1e-09} )$ & $\num{9.1e-01} \quad\hfill ( \pm\num{4e-01} )$ & $\num{8.1e-01} \quad\hfill ( \pm\num{3e-01} )$ \\
    & 80 & $\num{2.3e-09} \quad\hfill ( \pm\num{3e-10} )$ & $\num{2.8e-09} \quad\hfill ( \pm\num{2e-09} )$ & $\num{4.1e-01} \quad\hfill ( \pm\num{8e-02} )$ & $\num{7.9e-01} \quad\hfill ( \pm\num{5e-01} )$ \\
    & 96 & $\num{2.7e-09} \quad\hfill ( \pm\num{7e-10} )$ & $\num{3.2e-09} \quad\hfill ( \pm\num{1e-09} )$ & $\num{6.4e-01} \quad\hfill ( \pm\num{1e-01} )$ & $\num{6.8e-01} \quad\hfill ( \pm\num{2e-01} )$ \\
    \cline{1-6}
    \multirow[t]{11}{*}{Mott ring} & 2 & $\num{1.3e-06} \quad\hfill ( \pm\num{1e-07} )$ & $\num{6.9e-07} \quad\hfill ( \pm\num{2e-07} )$ & $\num{1.7e+02} \quad\hfill ( \pm\num{8e+01} )$ & $\num{7.4e+01} \quad\hfill ( \pm\num{2e+01} )$ \\
    & 3 & $\num{8.2e-07} \quad\hfill ( \pm\num{5e-07} )$ & $\num{6.3e-07} \quad\hfill ( \pm\num{3e-07} )$ & $\num{5.5e+01} \quad\hfill ( \pm\num{3e+01} )$ & $\num{2.5e+01} \quad\hfill ( \pm\num{2e+01} )$ \\
    & 4 & $\num{6.1e-07} \quad\hfill ( \pm\num{4e-07} )$ & $\num{5.1e-07} \quad\hfill ( \pm\num{4e-07} )$ & $\num{4.0e+01} \quad\hfill ( \pm\num{2e+01} )$ & $\num{3.0e+01} \quad\hfill ( \pm\num{1e+01} )$ \\
    & 8 & $\num{1.4e-07} \quad\hfill ( \pm\num{6e-08} )$ & $\num{6.4e-08} \quad\hfill ( \pm\num{5e-08} )$ & $\num{2.8e+01} \quad\hfill ( \pm\num{1e+01} )$ & $\num{9.5e+00} \quad\hfill ( \pm\num{3e+00} )$ \\
    & 16 & $\num{1.8e-08} \quad\hfill ( \pm\num{4e-09} )$ & $\num{1.5e-08} \quad\hfill ( \pm\num{2e-09} )$ & $\num{1.1e+01} \quad\hfill ( \pm\num{1e+01} )$ & $\num{2.0e+00} \quad\hfill ( \pm\num{2e-01} )$ \\
    & 24 & $\num{3.2e-08} \quad\hfill ( \pm\num{2e-08} )$ & $\num{3.4e-08} \quad\hfill ( \pm\num{3e-08} )$ & $\num{6.9e+00} \quad\hfill ( \pm\num{6e+00} )$ & $\num{4.3e+00} \quad\hfill ( \pm\num{3e+00} )$ \\
    & 32 & $\num{1.2e-08} \quad\hfill ( \pm\num{4e-09} )$ & $\num{5.7e-09} \quad\hfill ( \pm\num{9e-10} )$ & $\num{1.1e+01} \quad\hfill ( \pm\num{6e+00} )$ & $\num{1.2e+00} \quad\hfill ( \pm\num{2e-01} )$ \\
    & 48 & $\num{1.4e-08} \quad\hfill ( \pm\num{6e-09} )$ & $\num{4.0e-09} \quad\hfill ( \pm\num{1e-10} )$ & $\num{5.7e+00} \quad\hfill ( \pm\num{3e+00} )$ & $\num{7.2e-01} \quad\hfill ( \pm\num{7e-02} )$ \\
    & 64 & $\num{4.2e-09} \quad\hfill ( \pm\num{5e-10} )$ & $\num{2.7e-09} \quad\hfill ( \pm\num{8e-10} )$ & $\num{3.7e+00} \quad\hfill ( \pm\num{1e+00} )$ & $\num{1.4e+00} \quad\hfill ( \pm\num{1e+00} )$ \\
    & 80 & $\num{3.4e-09} \quad\hfill ( \pm\num{2e-09} )$ & $\num{1.9e-09} \quad\hfill ( \pm\num{5e-11} )$ & $\num{2.9e+00} \quad\hfill ( \pm\num{2e+00} )$ & $\num{4.1e-01} \quad\hfill ( \pm\num{2e-02} )$ \\
    & 96 & $\num{3.8e-09} \quad\hfill ( \pm\num{2e-09} )$ & $\num{3.5e-09} \quad\hfill ( \pm\num{2e-09} )$ & $\num{3.4e+00} \quad\hfill ( \pm\num{3e+00} )$ & $\num{1.0e+00} \quad\hfill ( \pm\num{9e-01} )$ \\
    \cline{1-6}
    \multirow[t]{11}{*}{Projectile} & 2 & $\num{1.3e-06} \quad\hfill ( \pm\num{9e-07} )$ & $\num{6.1e-07} \quad\hfill ( \pm\num{5e-07} )$ & $\num{4.5e+03} \quad\hfill ( \pm\num{2e+03} )$ & $\num{2.8e+03} \quad\hfill ( \pm\num{8e+02} )$ \\
    & 3 & $\num{2.1e-07} \quad\hfill ( \pm\num{6e-08} )$ & $\num{1.3e-07} \quad\hfill ( \pm\num{6e-09} )$ & $\num{2.2e+03} \quad\hfill ( \pm\num{2e+02} )$ & $\num{2.1e+03} \quad\hfill ( \pm\num{3e+02} )$ \\
    & 4 & $\num{1.3e-07} \quad\hfill ( \pm\num{4e-08} )$ & $\num{9.3e-08} \quad\hfill ( \pm\num{5e-09} )$ & $\num{1.5e+03} \quad\hfill ( \pm\num{9e+01} )$ & $\num{1.3e+03} \quad\hfill ( \pm\num{1e+01} )$ \\
    & 8 & $\num{3.8e-08} \quad\hfill ( \pm\num{1e-08} )$ & $\num{3.0e-08} \quad\hfill ( \pm\num{5e-09} )$ & $\num{1.1e+03} \quad\hfill ( \pm\num{5e+01} )$ & $\num{1.0e+03} \quad\hfill ( \pm\num{2e+01} )$ \\
    & 16 & $\num{2.7e-08} \quad\hfill ( \pm\num{1e-08} )$ & $\num{2.1e-08} \quad\hfill ( \pm\num{3e-09} )$ & $\num{1.0e+03} \quad\hfill ( \pm\num{1e+02} )$ & $\num{1.0e+03} \quad\hfill ( \pm\num{4e+01} )$ \\
    & 24 & $\num{4.9e-08} \quad\hfill ( \pm\num{2e-08} )$ & $\num{2.1e-08} \quad\hfill ( \pm\num{9e-10} )$ & $\num{1.1e+03} \quad\hfill ( \pm\num{1e+02} )$ & $\num{9.0e+02} \quad\hfill ( \pm\num{2e+01} )$ \\
    & 32 & $\num{1.5e-08} \quad\hfill ( \pm\num{1e-09} )$ & $\num{1.5e-08} \quad\hfill ( \pm\num{4e-09} )$ & $\num{6.5e+02} \quad\hfill ( \pm\num{2e+01} )$ & $\num{6.7e+02} \quad\hfill ( \pm\num{4e+01} )$ \\
    & 48 & $\num{1.4e-08} \quad\hfill ( \pm\num{3e-09} )$ & $\num{1.3e-08} \quad\hfill ( \pm\num{8e-10} )$ & $\num{6.2e+02} \quad\hfill ( \pm\num{5e+01} )$ & $\num{5.9e+02} \quad\hfill ( \pm\num{2e+01} )$ \\
    & 64 & $\num{1.8e-08} \quad\hfill ( \pm\num{5e-09} )$ & $\num{1.5e-08} \quad\hfill ( \pm\num{4e-09} )$ & $\num{7.0e+02} \quad\hfill ( \pm\num{1e+02} )$ & $\num{6.5e+02} \quad\hfill ( \pm\num{8e+01} )$ \\
    & 80 & $\num{3.6e-08} \quad\hfill ( \pm\num{1e-08} )$ & $\num{1.0e-08} \quad\hfill ( \pm\num{9e-10} )$ & $\num{7.5e+02} \quad\hfill ( \pm\num{7e+01} )$ & $\num{5.6e+02} \quad\hfill ( \pm\num{3e+01} )$ \\
    & 96 & $\num{3.0e-08} \quad\hfill ( \pm\num{2e-08} )$ & $\num{1.6e-08} \quad\hfill ( \pm\num{5e-09} )$ & $\num{7.8e+02} \quad\hfill ( \pm\num{1e+02} )$ & $\num{6.7e+02} \quad\hfill ( \pm\num{7e+01} )$ \\
    \cline{1-6}
    \bottomrule
  \end{tabular}
  \caption{
    Average and standard deviation of MSE ($\mu \pm \sigma$) on position prediction (MSE$_x$) and velocity prediction (MSE$_v$) across different datasets and number of basis functions on basis learning task.
    }
    \label{tb:basis_mse}
\end{table*}

\begin{table*}
  \begin{tabular}{
      |l
      |c
      |c
      |c
      |c
      |c|
    }
    \hline
    &  & Train MAE$_x$ & Test MAE$_x$ & Train MAE$_v$ & Test MAE$_v$ \\
    Dataset & Basis ($K$) &  &  &  &  \\
    \hline
    \multirow[t]{11}{*}{Exploding} & 2 & $\num{1.7e-03} \quad\hfill ( \pm\num{3e-04} )$ & $\num{2.7e-03} \quad\hfill ( \pm\num{3e-04} )$ & $\num{9.6e+00} \quad\hfill ( \pm\num{1e+00} )$ & $\num{1.2e+01} \quad\hfill ( \pm\num{1e+00} )$ \\
    & 3 & $\num{1.5e-03} \quad\hfill ( \pm\num{4e-04} )$ & $\num{1.7e-03} \quad\hfill ( \pm\num{5e-04} )$ & $\num{7.4e+00} \quad\hfill ( \pm\num{3e-01} )$ & $\num{7.5e+00} \quad\hfill ( \pm\num{6e-01} )$ \\
    & 4 & $\num{8.0e-04} \quad\hfill ( \pm\num{3e-04} )$ & $\num{8.8e-04} \quad\hfill ( \pm\num{5e-04} )$ & $\num{6.4e+00} \quad\hfill ( \pm\num{3e-01} )$ & $\num{6.3e+00} \quad\hfill ( \pm\num{4e-01} )$ \\
    & 8 & $\num{5.8e-04} \quad\hfill ( \pm\num{1e-04} )$ & $\num{5.9e-04} \quad\hfill ( \pm\num{8e-05} )$ & $\num{3.2e+00} \quad\hfill ( \pm\num{8e-02} )$ & $\num{3.1e+00} \quad\hfill ( \pm\num{2e-01} )$ \\
    & 16 & $\num{1.6e-04} \quad\hfill ( \pm\num{2e-05} )$ & $\num{1.1e-04} \quad\hfill ( \pm\num{3e-05} )$ & $\num{1.2e+00} \quad\hfill ( \pm\num{5e-02} )$ & $\num{9.9e-01} \quad\hfill ( \pm\num{3e-01} )$ \\
    & 24 & $\num{7.0e-05} \quad\hfill ( \pm\num{2e-06} )$ & $\num{7.0e-05} \quad\hfill ( \pm\num{5e-06} )$ & $\num{8.8e-01} \quad\hfill ( \pm\num{1e-01} )$ & $\num{7.6e-01} \quad\hfill ( \pm\num{1e-01} )$ \\
    & 32 & $\num{4.4e-05} \quad\hfill ( \pm\num{9e-06} )$ & $\num{4.0e-05} \quad\hfill ( \pm\num{5e-06} )$ & $\num{6.0e-01} \quad\hfill ( \pm\num{2e-01} )$ & $\num{5.4e-01} \quad\hfill ( \pm\num{2e-01} )$ \\
    & 48 & $\num{4.5e-05} \quad\hfill ( \pm\num{8e-06} )$ & $\num{4.0e-05} \quad\hfill ( \pm\num{5e-06} )$ & $\num{4.9e-01} \quad\hfill ( \pm\num{4e-02} )$ & $\num{5.8e-01} \quad\hfill ( \pm\num{1e-01} )$ \\
    & 64 & $\num{4.3e-05} \quad\hfill ( \pm\num{3e-06} )$ & $\num{4.4e-05} \quad\hfill ( \pm\num{3e-06} )$ & $\num{5.6e-01} \quad\hfill ( \pm\num{1e-01} )$ & $\num{5.7e-01} \quad\hfill ( \pm\num{1e-01} )$ \\
    & 80 & $\num{3.4e-05} \quad\hfill ( \pm\num{3e-06} )$ & $\num{3.5e-05} \quad\hfill ( \pm\num{1e-05} )$ & $\num{3.7e-01} \quad\hfill ( \pm\num{9e-02} )$ & $\num{4.9e-01} \quad\hfill ( \pm\num{2e-01} )$ \\
    & 96 & $\num{3.8e-05} \quad\hfill ( \pm\num{3e-06} )$ & $\num{3.9e-05} \quad\hfill ( \pm\num{6e-06} )$ & $\num{5.1e-01} \quad\hfill ( \pm\num{6e-02} )$ & $\num{5.0e-01} \quad\hfill ( \pm\num{8e-02} )$ \\
    \cline{1-6}
    \multirow[t]{11}{*}{Mott ring} & 2 & $\num{8.3e-04} \quad\hfill ( \pm\num{6e-05} )$ & $\num{6.1e-04} \quad\hfill ( \pm\num{6e-05} )$ & $\num{8.1e+00} \quad\hfill ( \pm\num{2e+00} )$ & $\num{5.1e+00} \quad\hfill ( \pm\num{5e-01} )$ \\
    & 3 & $\num{6.1e-04} \quad\hfill ( \pm\num{2e-04} )$ & $\num{5.5e-04} \quad\hfill ( \pm\num{1e-04} )$ & $\num{4.4e+00} \quad\hfill ( \pm\num{1e+00} )$ & $\num{3.2e+00} \quad\hfill ( \pm\num{7e-01} )$ \\
    & 4 & $\num{5.5e-04} \quad\hfill ( \pm\num{2e-04} )$ & $\num{4.8e-04} \quad\hfill ( \pm\num{2e-04} )$ & $\num{4.0e+00} \quad\hfill ( \pm\num{1e+00} )$ & $\num{3.4e+00} \quad\hfill ( \pm\num{9e-01} )$ \\
    & 8 & $\num{2.7e-04} \quad\hfill ( \pm\num{6e-05} )$ & $\num{1.8e-04} \quad\hfill ( \pm\num{7e-05} )$ & $\num{3.7e+00} \quad\hfill ( \pm\num{5e-01} )$ & $\num{2.2e+00} \quad\hfill ( \pm\num{3e-01} )$ \\
    & 16 & $\num{9.8e-05} \quad\hfill ( \pm\num{1e-05} )$ & $\num{9.2e-05} \quad\hfill ( \pm\num{8e-06} )$ & $\num{1.9e+00} \quad\hfill ( \pm\num{1e+00} )$ & $\num{9.7e-01} \quad\hfill ( \pm\num{4e-02} )$ \\
    & 24 & $\num{1.3e-04} \quad\hfill ( \pm\num{5e-05} )$ & $\num{1.3e-04} \quad\hfill ( \pm\num{5e-05} )$ & $\num{1.6e+00} \quad\hfill ( \pm\num{8e-01} )$ & $\num{1.4e+00} \quad\hfill ( \pm\num{6e-01} )$ \\
    & 32 & $\num{8.1e-05} \quad\hfill ( \pm\num{2e-05} )$ & $\num{5.6e-05} \quad\hfill ( \pm\num{6e-06} )$ & $\num{2.2e+00} \quad\hfill ( \pm\num{9e-01} )$ & $\num{7.1e-01} \quad\hfill ( \pm\num{3e-02} )$ \\
    & 48 & $\num{8.5e-05} \quad\hfill ( \pm\num{2e-05} )$ & $\num{4.6e-05} \quad\hfill ( \pm\num{4e-07} )$ & $\num{1.7e+00} \quad\hfill ( \pm\num{6e-01} )$ & $\num{5.8e-01} \quad\hfill ( \pm\num{1e-02} )$ \\
    & 64 & $\num{4.9e-05} \quad\hfill ( \pm\num{2e-06} )$ & $\num{3.8e-05} \quad\hfill ( \pm\num{7e-06} )$ & $\num{1.3e+00} \quad\hfill ( \pm\num{3e-01} )$ & $\num{7.7e-01} \quad\hfill ( \pm\num{4e-01} )$ \\
    & 80 & $\num{4.4e-05} \quad\hfill ( \pm\num{1e-05} )$ & $\num{3.3e-05} \quad\hfill ( \pm\num{7e-07} )$ & $\num{1.0e+00} \quad\hfill ( \pm\num{6e-01} )$ & $\num{4.7e-01} \quad\hfill ( \pm\num{1e-02} )$ \\
    & 96 & $\num{4.3e-05} \quad\hfill ( \pm\num{1e-05} )$ & $\num{4.2e-05} \quad\hfill ( \pm\num{1e-05} )$ & $\num{1.1e+00} \quad\hfill ( \pm\num{7e-01} )$ & $\num{6.8e-01} \quad\hfill ( \pm\num{3e-01} )$ \\
    \cline{1-6}
    \multirow[t]{11}{*}{Projectile} & 2 & $\num{5.1e-04} \quad\hfill ( \pm\num{1e-04} )$ & $\num{3.6e-04} \quad\hfill ( \pm\num{1e-04} )$ & $\num{3.1e+01} \quad\hfill ( \pm\num{1e+01} )$ & $\num{2.5e+01} \quad\hfill ( \pm\num{7e+00} )$ \\
    & 3 & $\num{2.2e-04} \quad\hfill ( \pm\num{3e-05} )$ & $\num{1.9e-04} \quad\hfill ( \pm\num{2e-05} )$ & $\num{1.9e+01} \quad\hfill ( \pm\num{2e+00} )$ & $\num{2.0e+01} \quad\hfill ( \pm\num{5e+00} )$ \\
    & 4 & $\num{1.9e-04} \quad\hfill ( \pm\num{2e-05} )$ & $\num{1.8e-04} \quad\hfill ( \pm\num{7e-06} )$ & $\num{1.7e+01} \quad\hfill ( \pm\num{2e+00} )$ & $\num{1.6e+01} \quad\hfill ( \pm\num{6e-01} )$ \\
    & 8 & $\num{9.3e-05} \quad\hfill ( \pm\num{5e-06} )$ & $\num{8.8e-05} \quad\hfill ( \pm\num{4e-06} )$ & $\num{1.4e+01} \quad\hfill ( \pm\num{2e-01} )$ & $\num{1.4e+01} \quad\hfill ( \pm\num{1e+00} )$ \\
    & 16 & $\num{8.3e-05} \quad\hfill ( \pm\num{1e-05} )$ & $\num{7.8e-05} \quad\hfill ( \pm\num{4e-06} )$ & $\num{1.4e+01} \quad\hfill ( \pm\num{8e-01} )$ & $\num{1.4e+01} \quad\hfill ( \pm\num{7e-01} )$ \\
    & 24 & $\num{1.0e-04} \quad\hfill ( \pm\num{1e-05} )$ & $\num{8.7e-05} \quad\hfill ( \pm\num{5e-06} )$ & $\num{1.3e+01} \quad\hfill ( \pm\num{7e-01} )$ & $\num{1.3e+01} \quad\hfill ( \pm\num{1e+00} )$ \\
    & 32 & $\num{7.2e-05} \quad\hfill ( \pm\num{2e-06} )$ & $\num{7.6e-05} \quad\hfill ( \pm\num{3e-06} )$ & $\num{1.1e+01} \quad\hfill ( \pm\num{1e-01} )$ & $\num{1.3e+01} \quad\hfill ( \pm\num{1e+00} )$ \\
    & 48 & $\num{6.5e-05} \quad\hfill ( \pm\num{4e-06} )$ & $\num{6.8e-05} \quad\hfill ( \pm\num{2e-06} )$ & $\num{1.2e+01} \quad\hfill ( \pm\num{4e-01} )$ & $\num{1.2e+01} \quad\hfill ( \pm\num{9e-01} )$ \\
    & 64 & $\num{8.0e-05} \quad\hfill ( \pm\num{2e-05} )$ & $\num{6.4e-05} \quad\hfill ( \pm\num{2e-06} )$ & $\num{1.3e+01} \quad\hfill ( \pm\num{2e+00} )$ & $\num{1.1e+01} \quad\hfill ( \pm\num{3e-01} )$ \\
    & 80 & $\num{8.9e-05} \quad\hfill ( \pm\num{2e-05} )$ & $\num{5.9e-05} \quad\hfill ( \pm\num{2e-06} )$ & $\num{1.2e+01} \quad\hfill ( \pm\num{5e-01} )$ & $\num{1.1e+01} \quad\hfill ( \pm\num{2e-01} )$ \\
    & 96 & $\num{8.6e-05} \quad\hfill ( \pm\num{5e-06} )$ & $\num{6.9e-05} \quad\hfill ( \pm\num{5e-06} )$ & $\num{1.3e+01} \quad\hfill ( \pm\num{6e-01} )$ & $\num{1.1e+01} \quad\hfill ( \pm\num{8e-01} )$ \\
    \cline{1-6}
    \bottomrule
  \end{tabular}
  \caption{
    Average and standard deviation of MAE ($\mu \pm \sigma$) on position prediction (MAE$_x$) and velocity prediction (MAE$_v$) across different datasets and number of basis functions on basis learning task.
  }
  \label{tb:basis_mae}
\end{table*}

\subsection{Dynamic}
\begin{table*}
  \begin{tabular}{
      |l
      |c
      |c
      |c|
    }
    \hline
    &  & Train R$^2$ & Test R$^2$ \\
    Dataset & Basis ($K$) &  &  \\
    \hline
    \multirow[t]{11}{*}{Exploding} & 2 & $\num{98.1098} \quad\hfill ( \pm \num{0.1578} )$ & $\num{98.5908} \quad\hfill ( \pm \num{0.0866} )$ \\
    & 3 & $\num{98.6320} \quad\hfill ( \pm \num{0.3910} )$ & $\num{99.0251} \quad\hfill ( \pm \num{0.0553} )$ \\
    & 4 & $\num{98.8048} \quad\hfill ( \pm \num{0.3080} )$ & $\num{99.3325} \quad\hfill ( \pm \num{0.0323} )$ \\
    & 8 & $\num{99.7885} \quad\hfill ( \pm \num{0.0552} )$ & $\num{99.8876} \quad\hfill ( \pm \num{0.0054} )$ \\
    & 16 & $\num{99.9890} \quad\hfill ( \pm \num{0.0083} )$ & $\num{99.9958} \quad\hfill ( \pm \num{0.0012} )$ \\
    & 24 & $\num{99.9948} \quad\hfill ( \pm \num{0.0020} )$ & $\num{99.9970} \quad\hfill ( \pm \num{0.0003} )$ \\
    & 32 & $\num{99.9982} \quad\hfill ( \pm \num{0.0002} )$ & $\num{99.9982} \quad\hfill ( \pm \num{0.0002} )$ \\
    & 48 & $\num{99.9981} \quad\hfill ( \pm \num{0.0005} )$ & $\num{99.9988} \quad\hfill ( \pm \num{0.0001} )$ \\
    & 64 & $\num{99.9980} \quad\hfill ( \pm \num{0.0006} )$ & $\num{99.9983} \quad\hfill ( \pm \num{0.0001} )$ \\
    & 80 & $\num{99.9983} \quad\hfill ( \pm \num{0.0002} )$ & $\num{99.9987} \quad\hfill ( \pm \num{0.0002} )$ \\
    & 96 & $\num{99.9984} \quad\hfill ( \pm \num{0.0005} )$ & $\num{99.9989} \quad\hfill ( \pm \num{0.0001} )$ \\
    \cline{1-4}
    \cline{1-4}
    \multirow[t]{11}{*}{Mott ring} & 2 & $\num{99.8808} \quad\hfill ( \pm \num{0.0116} )$ & $\num{99.9034} \quad\hfill ( \pm \num{0.0194} )$ \\
    & 3 & $\num{99.9382} \quad\hfill ( \pm \num{0.0303} )$ & $\num{99.9655} \quad\hfill ( \pm \num{0.0173} )$ \\
    & 4 & $\num{99.9436} \quad\hfill ( \pm \num{0.0364} )$ & $\num{99.9827} \quad\hfill ( \pm \num{0.0106} )$ \\
    & 8 & $\num{99.9699} \quad\hfill ( \pm \num{0.0170} )$ & $\num{99.9888} \quad\hfill ( \pm \num{0.0048} )$ \\
    & 16 & $\num{99.9849} \quad\hfill ( \pm \num{0.0121} )$ & $\num{99.9972} \quad\hfill ( \pm \num{0.0002} )$ \\
    & 24 & $\num{99.9891} \quad\hfill ( \pm \num{0.0041} )$ & $\num{99.9945} \quad\hfill ( \pm \num{0.0022} )$ \\
    & 32 & $\num{99.9941} \quad\hfill ( \pm \num{0.0049} )$ & $\num{99.9971} \quad\hfill ( \pm \num{0.0022} )$ \\
    & 48 & $\num{99.9912} \quad\hfill ( \pm \num{0.0065} )$ & $\num{99.9969} \quad\hfill ( \pm \num{0.0021} )$ \\
    & 64 & $\num{99.9968} \quad\hfill ( \pm \num{0.0014} )$ & $\num{99.9979} \quad\hfill ( \pm \num{0.0013} )$ \\
    & 80 & $\num{99.9959} \quad\hfill ( \pm \num{0.0021} )$ & $\num{99.9981} \quad\hfill ( \pm \num{0.0011} )$ \\
    & 96 & $\num{99.9973} \quad\hfill ( \pm \num{0.0021} )$ & $\num{99.9992} \quad\hfill ( \pm \num{0.0002} )$ \\
    \cline{1-4}
    \multirow[t]{11}{*}{Projectile} & 2 & $\num{89.6531} \quad\hfill ( \pm \num{5.2606} )$ & $\num{94.8480} \quad\hfill ( \pm \num{0.5477} )$ \\
    & 3 & $\num{96.1619} \quad\hfill ( \pm \num{0.5678} )$ & $\num{96.7571} \quad\hfill ( \pm \num{0.1743} )$ \\
    & 4 & $\num{97.2062} \quad\hfill ( \pm \num{0.7094} )$ & $\num{97.7648} \quad\hfill ( \pm \num{0.0890} )$ \\
    & 8 & $\num{97.7770} \quad\hfill ( \pm \num{0.2725} )$ & $\num{98.6154} \quad\hfill ( \pm \num{0.0234} )$ \\
    & 16 & $\num{98.5947} \quad\hfill ( \pm \num{0.1866} )$ & $\num{98.6641} \quad\hfill ( \pm \num{0.1498} )$ \\
    & 24 & $\num{97.9801} \quad\hfill ( \pm \num{0.4938} )$ & $\num{98.9598} \quad\hfill ( \pm \num{0.0851} )$ \\
    & 32 & $\num{98.9991} \quad\hfill ( \pm \num{0.1804} )$ & $\num{99.0932} \quad\hfill ( \pm \num{0.0822} )$ \\
    & 48 & $\num{98.9858} \quad\hfill ( \pm \num{0.1579} )$ & $\num{99.1680} \quad\hfill ( \pm \num{0.0957} )$ \\
    & 64 & $\num{98.8969} \quad\hfill ( \pm \num{0.4136} )$ & $\num{99.0991} \quad\hfill ( \pm \num{0.2358} )$ \\
    & 80 & $\num{99.1490} \quad\hfill ( \pm \num{0.1599} )$ & $\num{99.2307} \quad\hfill ( \pm \num{0.0749} )$ \\
    & 96 & $\num{98.9779} \quad\hfill ( \pm \num{0.2903} )$ & $\num{99.0927} \quad\hfill ( \pm \num{0.1057} )$ \\
    \cline{1-4}
    \bottomrule
  \end{tabular}
  \caption{Average and standard deviation of R$^2$ ($\mu \pm \sigma$) across different datasets and number of basis functions on dynamic learning task.}
  \label{tb:full_r2_basis}
\end{table*}

\begin{table*}
  \begin{tabular}{
      |l
      |c
      |c
      |c
      |c
      |c|
    }
    \hline
    &  & Train MSE$_x$ & Test MSE$_x$ & Train MSE$_v$ & Test MSE$_v$ \\
    Dataset & Basis ($K$) &  &  &  &  \\
    \hline
    \multirow[t]{11}{*}{Exploding} & 2 & $\num{6.6e-06} \quad\hfill ( \pm\num{2e-06} )$ & $\num{1.8e-05} \quad\hfill ( \pm\num{5e-06} )$ & $\num{3.3e+02} \quad\hfill ( \pm\num{9e+01} )$ & $\num{2.8e+02} \quad\hfill ( \pm\num{4e+01} )$ \\
    & 3 & $\num{9.8e-06} \quad\hfill ( \pm\num{3e-06} )$ & $\num{1.2e-05} \quad\hfill ( \pm\num{6e-06} )$ & $\num{1.7e+02} \quad\hfill ( \pm\num{5e+00} )$ & $\num{1.8e+02} \quad\hfill ( \pm\num{6e+00} )$ \\
    & 4 & $\num{5.2e-06} \quad\hfill ( \pm\num{3e-06} )$ & $\num{7.4e-06} \quad\hfill ( \pm\num{4e-06} )$ & $\num{1.3e+02} \quad\hfill ( \pm\num{5e+00} )$ & $\num{1.3e+02} \quad\hfill ( \pm\num{5e+00} )$ \\
    & 8 & $\num{6.2e-07} \quad\hfill ( \pm\num{2e-07} )$ & $\num{1.4e-06} \quad\hfill ( \pm\num{3e-07} )$ & $\num{4.2e+01} \quad\hfill ( \pm\num{8e+00} )$ & $\num{3.0e+01} \quad\hfill ( \pm\num{1e+00} )$ \\
    & 16 & $\num{8.7e-08} \quad\hfill ( \pm\num{4e-08} )$ & $\num{3.2e-08} \quad\hfill ( \pm\num{1e-08} )$ & $\num{4.2e+00} \quad\hfill ( \pm\num{1e+00} )$ & $\num{2.0e+00} \quad\hfill ( \pm\num{7e-01} )$ \\
    & 24 & $\num{2.8e-08} \quad\hfill ( \pm\num{2e-08} )$ & $\num{4.4e-08} \quad\hfill ( \pm\num{2e-08} )$ & $\num{2.4e+00} \quad\hfill ( \pm\num{8e-01} )$ & $\num{2.3e+00} \quad\hfill ( \pm\num{6e-01} )$ \\
    & 32 & $\num{9.9e-09} \quad\hfill ( \pm\num{5e-09} )$ & $\num{1.0e-08} \quad\hfill ( \pm\num{5e-09} )$ & $\num{9.3e-01} \quad\hfill ( \pm\num{3e-01} )$ & $\num{9.3e-01} \quad\hfill ( \pm\num{4e-01} )$ \\
    & 48 & $\num{1.1e-08} \quad\hfill ( \pm\num{4e-09} )$ & $\num{1.8e-08} \quad\hfill ( \pm\num{3e-09} )$ & $\num{8.5e-01} \quad\hfill ( \pm\num{2e-01} )$ & $\num{1.0e+00} \quad\hfill ( \pm\num{1e-01} )$ \\
    & 64 & $\num{1.2e-08} \quad\hfill ( \pm\num{1e-09} )$ & $\num{6.9e-09} \quad\hfill ( \pm\num{3e-09} )$ & $\num{7.7e-01} \quad\hfill ( \pm\num{1e-01} )$ & $\num{7.8e-01} \quad\hfill ( \pm\num{4e-01} )$ \\
    & 80 & $\num{1.7e-08} \quad\hfill ( \pm\num{1e-08} )$ & $\num{1.3e-08} \quad\hfill ( \pm\num{6e-09} )$ & $\num{9.7e-01} \quad\hfill ( \pm\num{6e-01} )$ & $\num{1.0e+00} \quad\hfill ( \pm\num{4e-01} )$ \\
    & 96 & $\num{1.2e-08} \quad\hfill ( \pm\num{2e-09} )$ & $\num{8.4e-09} \quad\hfill ( \pm\num{4e-09} )$ & $\num{8.9e-01} \quad\hfill ( \pm\num{2e-01} )$ & $\num{7.6e-01} \quad\hfill ( \pm\num{2e-01} )$ \\
    \cline{1-6}
    \multirow[t]{11}{*}{Mott ring} & 2 & $\num{1.0e-06} \quad\hfill ( \pm\num{4e-07} )$ & $\num{1.1e-06} \quad\hfill ( \pm\num{6e-07} )$ & $\num{1.1e+02} \quad\hfill ( \pm\num{2e+01} )$ & $\num{8.5e+01} \quad\hfill ( \pm\num{4e+01} )$ \\
    & 3 & $\num{5.0e-07} \quad\hfill ( \pm\num{2e-07} )$ & $\num{4.1e-07} \quad\hfill ( \pm\num{1e-07} )$ & $\num{5.6e+01} \quad\hfill ( \pm\num{3e+01} )$ & $\num{2.6e+01} \quad\hfill ( \pm\num{2e+01} )$ \\
    & 4 & $\num{5.7e-07} \quad\hfill ( \pm\num{4e-07} )$ & $\num{3.3e-07} \quad\hfill ( \pm\num{3e-07} )$ & $\num{4.4e+01} \quad\hfill ( \pm\num{2e+01} )$ & $\num{1.3e+01} \quad\hfill ( \pm\num{1e+01} )$ \\
    & 8 & $\num{1.6e-07} \quad\hfill ( \pm\num{1e-07} )$ & $\num{8.9e-08} \quad\hfill ( \pm\num{3e-08} )$ & $\num{2.5e+01} \quad\hfill ( \pm\num{1e+01} )$ & $\num{8.1e+00} \quad\hfill ( \pm\num{4e+00} )$ \\
    & 16 & $\num{3.3e-08} \quad\hfill ( \pm\num{2e-08} )$ & $\num{2.2e-08} \quad\hfill ( \pm\num{1e-08} )$ & $\num{1.4e+01} \quad\hfill ( \pm\num{1e+01} )$ & $\num{1.9e+00} \quad\hfill ( \pm\num{3e-01} )$ \\
    & 24 & $\num{4.6e-08} \quad\hfill ( \pm\num{2e-08} )$ & $\num{2.2e-08} \quad\hfill ( \pm\num{7e-09} )$ & $\num{1.2e+01} \quad\hfill ( \pm\num{4e+00} )$ & $\num{5.8e+00} \quad\hfill ( \pm\num{3e+00} )$ \\
    & 32 & $\num{2.1e-08} \quad\hfill ( \pm\num{2e-08} )$ & $\num{8.7e-09} \quad\hfill ( \pm\num{3e-09} )$ & $\num{5.9e+00} \quad\hfill ( \pm\num{5e+00} )$ & $\num{2.9e+00} \quad\hfill ( \pm\num{3e+00} )$ \\
    & 48 & $\num{2.0e-08} \quad\hfill ( \pm\num{8e-09} )$ & $\num{9.3e-09} \quad\hfill ( \pm\num{4e-09} )$ & $\num{7.6e+00} \quad\hfill ( \pm\num{5e+00} )$ & $\num{2.8e+00} \quad\hfill ( \pm\num{2e+00} )$ \\
    & 64 & $\num{6.6e-09} \quad\hfill ( \pm\num{3e-09} )$ & $\num{4.1e-09} \quad\hfill ( \pm\num{1e-09} )$ & $\num{3.9e+00} \quad\hfill ( \pm\num{2e+00} )$ & $\num{2.4e+00} \quad\hfill ( \pm\num{2e+00} )$ \\
    & 80 & $\num{1.1e-08} \quad\hfill ( \pm\num{1e-09} )$ & $\num{9.4e-09} \quad\hfill ( \pm\num{3e-09} )$ & $\num{4.2e+00} \quad\hfill ( \pm\num{2e+00} )$ & $\num{2.2e+00} \quad\hfill ( \pm\num{2e+00} )$ \\
    & 96 & $\num{7.1e-09} \quad\hfill ( \pm\num{2e-09} )$ & $\num{5.3e-09} \quad\hfill ( \pm\num{3e-10} )$ & $\num{2.7e+00} \quad\hfill ( \pm\num{2e+00} )$ & $\num{6.0e-01} \quad\hfill ( \pm\num{2e-01} )$ \\
    \cline{1-6}
    \multirow[t]{11}{*}{Projectile} & 2 & $\num{1.3e-06} \quad\hfill ( \pm\num{9e-07} )$ & $\num{4.1e-07} \quad\hfill ( \pm\num{3e-07} )$ & $\num{6.2e+03} \quad\hfill ( \pm\num{4e+03} )$ & $\num{2.5e+03} \quad\hfill ( \pm\num{4e+02} )$ \\
    & 3 & $\num{1.6e-07} \quad\hfill ( \pm\num{3e-08} )$ & $\num{1.6e-07} \quad\hfill ( \pm\num{3e-08} )$ & $\num{2.6e+03} \quad\hfill ( \pm\num{9e+01} )$ & $\num{2.1e+03} \quad\hfill ( \pm\num{2e+02} )$ \\
    & 4 & $\num{1.8e-07} \quad\hfill ( \pm\num{1e-07} )$ & $\num{1.0e-07} \quad\hfill ( \pm\num{1e-08} )$ & $\num{1.6e+03} \quad\hfill ( \pm\num{2e+02} )$ & $\num{1.4e+03} \quad\hfill ( \pm\num{3e+01} )$ \\
    & 8 & $\num{1.1e-07} \quad\hfill ( \pm\num{2e-08} )$ & $\num{2.9e-08} \quad\hfill ( \pm\num{5e-09} )$ & $\num{1.3e+03} \quad\hfill ( \pm\num{8e+01} )$ & $\num{1.0e+03} \quad\hfill ( \pm\num{3e+01} )$ \\
    & 16 & $\num{3.4e-08} \quad\hfill ( \pm\num{2e-08} )$ & $\num{2.6e-08} \quad\hfill ( \pm\num{1e-08} )$ & $\num{1.1e+03} \quad\hfill ( \pm\num{9e+01} )$ & $\num{1.1e+03} \quad\hfill ( \pm\num{7e+01} )$ \\
    & 24 & $\num{4.9e-08} \quad\hfill ( \pm\num{1e-08} )$ & $\num{2.1e-08} \quad\hfill ( \pm\num{3e-09} )$ & $\num{1.2e+03} \quad\hfill ( \pm\num{1e+02} )$ & $\num{9.0e+02} \quad\hfill ( \pm\num{3e+01} )$ \\
    & 32 & $\num{2.0e-08} \quad\hfill ( \pm\num{9e-09} )$ & $\num{1.2e-08} \quad\hfill ( \pm\num{2e-09} )$ & $\num{6.8e+02} \quad\hfill ( \pm\num{7e+01} )$ & $\num{6.4e+02} \quad\hfill ( \pm\num{2e+01} )$ \\
    & 48 & $\num{2.1e-08} \quad\hfill ( \pm\num{8e-09} )$ & $\num{1.3e-08} \quad\hfill ( \pm\num{1e-09} )$ & $\num{6.7e+02} \quad\hfill ( \pm\num{7e+01} )$ & $\num{6.0e+02} \quad\hfill ( \pm\num{4e+01} )$ \\
    & 64 & $\num{2.9e-08} \quad\hfill ( \pm\num{2e-08} )$ & $\num{1.6e-08} \quad\hfill ( \pm\num{7e-09} )$ & $\num{7.4e+02} \quad\hfill ( \pm\num{1e+02} )$ & $\num{6.2e+02} \quad\hfill ( \pm\num{6e+01} )$ \\
    & 80 & $\num{1.2e-08} \quad\hfill ( \pm\num{4e-09} )$ & $\num{1.1e-08} \quad\hfill ( \pm\num{2e-09} )$ & $\num{5.9e+02} \quad\hfill ( \pm\num{5e+01} )$ & $\num{5.7e+02} \quad\hfill ( \pm\num{4e+01} )$ \\
    & 96 & $\num{2.4e-08} \quad\hfill ( \pm\num{1e-08} )$ & $\num{1.9e-08} \quad\hfill ( \pm\num{6e-09} )$ & $\num{7.4e+02} \quad\hfill ( \pm\num{1e+02} )$ & $\num{7.1e+02} \quad\hfill ( \pm\num{6e+01} )$ \\
    \cline{1-6}
    \bottomrule
  \end{tabular}
  \caption{
    Average and standard deviation of MSE ($\mu \pm \sigma$) on position prediction (MSE$_x$) and velocity prediction (MSE$_v$) across different datasets and number of basis functions on dynamic learning task.
  }
  \label{tb:dynamic_mse}
\end{table*}

\begin{table*}
  \begin{tabular}{
      |l
      |c
      |c
      |c
      |c
      |c|
    }
    \hline
    &  & Train MAE$_x$ & Test MAE$_x$ & Train MAE$_v$ & Test MAE$_v$ \\
    Dataset & Basis ($K$) &  &  &  &  \\
    \hline
    \multirow[t]{11}{*}{Exploding} & 2 & $\num{1.5e-03} \quad\hfill ( \pm\num{1e-04} )$ & $\num{2.2e-03} \quad\hfill ( \pm\num{4e-04} )$ & $\num{1.0e+01} \quad\hfill ( \pm\num{2e+00} )$ & $\num{9.7e+00} \quad\hfill ( \pm\num{1e+00} )$ \\
    & 3 & $\num{1.5e-03} \quad\hfill ( \pm\num{2e-04} )$ & $\num{1.7e-03} \quad\hfill ( \pm\num{5e-04} )$ & $\num{7.2e+00} \quad\hfill ( \pm\num{4e-01} )$ & $\num{7.6e+00} \quad\hfill ( \pm\num{6e-01} )$ \\
    & 4 & $\num{1.1e-03} \quad\hfill ( \pm\num{3e-04} )$ & $\num{1.3e-03} \quad\hfill ( \pm\num{3e-04} )$ & $\num{5.8e+00} \quad\hfill ( \pm\num{3e-01} )$ & $\num{6.1e+00} \quad\hfill ( \pm\num{5e-01} )$ \\
    & 8 & $\num{4.4e-04} \quad\hfill ( \pm\num{2e-05} )$ & $\num{5.7e-04} \quad\hfill ( \pm\num{6e-05} )$ & $\num{3.3e+00} \quad\hfill ( \pm\num{3e-01} )$ & $\num{3.1e+00} \quad\hfill ( \pm\num{1e-01} )$ \\
    & 16 & $\num{1.6e-04} \quad\hfill ( \pm\num{4e-05} )$ & $\num{9.2e-05} \quad\hfill ( \pm\num{2e-05} )$ & $\num{1.3e+00} \quad\hfill ( \pm\num{2e-01} )$ & $\num{8.8e-01} \quad\hfill ( \pm\num{2e-01} )$ \\
    & 24 & $\num{9.8e-05} \quad\hfill ( \pm\num{2e-05} )$ & $\num{1.0e-04} \quad\hfill ( \pm\num{2e-05} )$ & $\num{9.4e-01} \quad\hfill ( \pm\num{2e-01} )$ & $\num{8.5e-01} \quad\hfill ( \pm\num{1e-01} )$ \\
    & 32 & $\num{7.0e-05} \quad\hfill ( \pm\num{2e-05} )$ & $\num{7.2e-05} \quad\hfill ( \pm\num{2e-05} )$ & $\num{6.6e-01} \quad\hfill ( \pm\num{2e-01} )$ & $\num{6.4e-01} \quad\hfill ( \pm\num{2e-01} )$ \\
    & 48 & $\num{8.0e-05} \quad\hfill ( \pm\num{1e-05} )$ & $\num{9.5e-05} \quad\hfill ( \pm\num{1e-05} )$ & $\num{6.2e-01} \quad\hfill ( \pm\num{1e-01} )$ & $\num{6.9e-01} \quad\hfill ( \pm\num{4e-02} )$ \\
    & 64 & $\num{8.3e-05} \quad\hfill ( \pm\num{1e-05} )$ & $\num{6.0e-05} \quad\hfill ( \pm\num{1e-05} )$ & $\num{5.9e-01} \quad\hfill ( \pm\num{5e-02} )$ & $\num{5.6e-01} \quad\hfill ( \pm\num{2e-01} )$ \\
    & 80 & $\num{9.5e-05} \quad\hfill ( \pm\num{4e-05} )$ & $\num{8.2e-05} \quad\hfill ( \pm\num{2e-05} )$ & $\num{5.7e-01} \quad\hfill ( \pm\num{2e-01} )$ & $\num{6.6e-01} \quad\hfill ( \pm\num{2e-01} )$ \\
    & 96 & $\num{8.0e-05} \quad\hfill ( \pm\num{8e-06} )$ & $\num{6.2e-05} \quad\hfill ( \pm\num{2e-05} )$ & $\num{5.9e-01} \quad\hfill ( \pm\num{1e-01} )$ & $\num{5.5e-01} \quad\hfill ( \pm\num{1e-01} )$ \\
    \cline{1-6}
    \multirow[t]{11}{*}{Mott ring} & 2 & $\num{7.0e-04} \quad\hfill ( \pm\num{1e-04} )$ & $\num{7.3e-04} \quad\hfill ( \pm\num{2e-04} )$ & $\num{6.1e+00} \quad\hfill ( \pm\num{2e-01} )$ & $\num{5.7e+00} \quad\hfill ( \pm\num{1e+00} )$ \\
    & 3 & $\num{5.0e-04} \quad\hfill ( \pm\num{1e-04} )$ & $\num{4.7e-04} \quad\hfill ( \pm\num{7e-05} )$ & $\num{4.3e+00} \quad\hfill ( \pm\num{1e+00} )$ & $\num{3.2e+00} \quad\hfill ( \pm\num{9e-01} )$ \\
    & 4 & $\num{5.3e-04} \quad\hfill ( \pm\num{2e-04} )$ & $\num{3.9e-04} \quad\hfill ( \pm\num{2e-04} )$ & $\num{4.1e+00} \quad\hfill ( \pm\num{1e+00} )$ & $\num{2.4e+00} \quad\hfill ( \pm\num{9e-01} )$ \\
    & 8 & $\num{2.8e-04} \quad\hfill ( \pm\num{1e-04} )$ & $\num{2.2e-04} \quad\hfill ( \pm\num{4e-05} )$ & $\num{3.3e+00} \quad\hfill ( \pm\num{9e-01} )$ & $\num{2.1e+00} \quad\hfill ( \pm\num{6e-01} )$ \\
    & 16 & $\num{1.3e-04} \quad\hfill ( \pm\num{4e-05} )$ & $\num{1.1e-04} \quad\hfill ( \pm\num{2e-05} )$ & $\num{2.4e+00} \quad\hfill ( \pm\num{1e+00} )$ & $\num{1.0e+00} \quad\hfill ( \pm\num{8e-02} )$ \\
    & 24 & $\num{1.7e-04} \quad\hfill ( \pm\num{4e-05} )$ & $\num{1.1e-04} \quad\hfill ( \pm\num{2e-05} )$ & $\num{2.4e+00} \quad\hfill ( \pm\num{4e-01} )$ & $\num{1.7e+00} \quad\hfill ( \pm\num{5e-01} )$ \\
    & 32 & $\num{1.0e-04} \quad\hfill ( \pm\num{4e-05} )$ & $\num{6.9e-05} \quad\hfill ( \pm\num{1e-05} )$ & $\num{1.5e+00} \quad\hfill ( \pm\num{8e-01} )$ & $\num{1.1e+00} \quad\hfill ( \pm\num{5e-01} )$ \\
    & 48 & $\num{1.0e-04} \quad\hfill ( \pm\num{3e-05} )$ & $\num{6.9e-05} \quad\hfill ( \pm\num{1e-05} )$ & $\num{1.8e+00} \quad\hfill ( \pm\num{8e-01} )$ & $\num{1.0e+00} \quad\hfill ( \pm\num{4e-01} )$ \\
    & 64 & $\num{6.0e-05} \quad\hfill ( \pm\num{2e-05} )$ & $\num{4.8e-05} \quad\hfill ( \pm\num{8e-06} )$ & $\num{1.3e+00} \quad\hfill ( \pm\num{5e-01} )$ & $\num{1.0e+00} \quad\hfill ( \pm\num{4e-01} )$ \\
    & 80 & $\num{7.9e-05} \quad\hfill ( \pm\num{4e-06} )$ & $\num{7.2e-05} \quad\hfill ( \pm\num{1e-05} )$ & $\num{1.4e+00} \quad\hfill ( \pm\num{3e-01} )$ & $\num{9.6e-01} \quad\hfill ( \pm\num{4e-01} )$ \\
    & 96 & $\num{6.2e-05} \quad\hfill ( \pm\num{1e-05} )$ & $\num{5.5e-05} \quad\hfill ( \pm\num{1e-06} )$ & $\num{1.0e+00} \quad\hfill ( \pm\num{5e-01} )$ & $\num{5.4e-01} \quad\hfill ( \pm\num{4e-02} )$ \\
    \cline{1-6}
    \multirow[t]{11}{*}{Projectile} & 2 & $\num{6.0e-04} \quad\hfill ( \pm\num{3e-04} )$ & $\num{3.2e-04} \quad\hfill ( \pm\num{8e-05} )$ & $\num{3.7e+01} \quad\hfill ( \pm\num{2e+01} )$ & $\num{2.3e+01} \quad\hfill ( \pm\num{4e+00} )$ \\
    & 3 & $\num{2.2e-04} \quad\hfill ( \pm\num{9e-06} )$ & $\num{2.0e-04} \quad\hfill ( \pm\num{2e-05} )$ & $\num{2.7e+01} \quad\hfill ( \pm\num{2e+00} )$ & $\num{2.0e+01} \quad\hfill ( \pm\num{3e+00} )$ \\
    & 4 & $\num{2.1e-04} \quad\hfill ( \pm\num{4e-05} )$ & $\num{1.8e-04} \quad\hfill ( \pm\num{6e-06} )$ & $\num{1.7e+01} \quad\hfill ( \pm\num{2e+00} )$ & $\num{1.5e+01} \quad\hfill ( \pm\num{6e-01} )$ \\
    & 8 & $\num{1.3e-04} \quad\hfill ( \pm\num{1e-05} )$ & $\num{8.7e-05} \quad\hfill ( \pm\num{4e-06} )$ & $\num{1.6e+01} \quad\hfill ( \pm\num{8e-01} )$ & $\num{1.4e+01} \quad\hfill ( \pm\num{1e+00} )$ \\
    & 16 & $\num{8.7e-05} \quad\hfill ( \pm\num{1e-05} )$ & $\num{8.1e-05} \quad\hfill ( \pm\num{7e-06} )$ & $\num{1.4e+01} \quad\hfill ( \pm\num{3e-01} )$ & $\num{1.4e+01} \quad\hfill ( \pm\num{8e-01} )$ \\
    & 24 & $\num{1.2e-04} \quad\hfill ( \pm\num{3e-05} )$ & $\num{7.9e-05} \quad\hfill ( \pm\num{4e-06} )$ & $\num{1.7e+01} \quad\hfill ( \pm\num{3e+00} )$ & $\num{1.3e+01} \quad\hfill ( \pm\num{3e-01} )$ \\
    & 32 & $\num{7.4e-05} \quad\hfill ( \pm\num{1e-05} )$ & $\num{6.8e-05} \quad\hfill ( \pm\num{3e-06} )$ & $\num{1.1e+01} \quad\hfill ( \pm\num{3e-01} )$ & $\num{1.3e+01} \quad\hfill ( \pm\num{2e+00} )$ \\
    & 48 & $\num{6.7e-05} \quad\hfill ( \pm\num{7e-06} )$ & $\num{5.9e-05} \quad\hfill ( \pm\num{1e-06} )$ & $\num{1.1e+01} \quad\hfill ( \pm\num{2e-01} )$ & $\num{1.2e+01} \quad\hfill ( \pm\num{1e+00} )$ \\
    & 64 & $\num{7.6e-05} \quad\hfill ( \pm\num{2e-05} )$ & $\num{7.7e-05} \quad\hfill ( \pm\num{2e-05} )$ & $\num{1.1e+01} \quad\hfill ( \pm\num{1e+00} )$ & $\num{1.2e+01} \quad\hfill ( \pm\num{1e+00} )$ \\
    & 80 & $\num{6.0e-05} \quad\hfill ( \pm\num{6e-06} )$ & $\num{5.7e-05} \quad\hfill ( \pm\num{2e-06} )$ & $\num{1.2e+01} \quad\hfill ( \pm\num{1e+00} )$ & $\num{1.1e+01} \quad\hfill ( \pm\num{6e-01} )$ \\
    & 96 & $\num{7.5e-05} \quad\hfill ( \pm\num{1e-05} )$ & $\num{6.8e-05} \quad\hfill ( \pm\num{3e-06} )$ & $\num{1.2e+01} \quad\hfill ( \pm\num{1e+00} )$ & $\num{1.2e+01} \quad\hfill ( \pm\num{9e-01} )$ \\
    \cline{1-6}
    \bottomrule
  \end{tabular}
  \caption{
    Average and standard deviation of MAE ($\mu \pm \sigma$) on position prediction (MAE$_x$) and velocity prediction (MAE$_v$) across different datasets and number of basis functions on dynamic learning task.
  }
  \label{tb:dynamic_mae}
\end{table*}

\end{document}